\documentclass[10pt,twocolumn,letterpaper]{article}
\pdfoutput=1

\usepackage[pagenumbers]{iccv} 

\usepackage{graphicx}
\usepackage{subcaption} 
\usepackage{amsmath}
\usepackage{amssymb}

\usepackage{todonotes}

\usepackage{multirow}
\usepackage{booktabs}
\usepackage{ragged2e}
\usepackage{algorithm}
\usepackage{algpseudocode}
\usepackage{microtype}
\usepackage{array}

\usepackage{color, colortbl}

\usepackage{pifont} 

\usepackage{xcolor}  
\definecolor{darkgreen}{rgb}{0.0, 0.5, 0.0}  

\definecolor{darkbrown}{rgb}{0.55, 0.14, 0.07} 

\newcommand{\up}[1]{\textcolor{darkgreen}{\tiny ($\uparrow$ #1)}}
\newcommand{\down}[1]{\textcolor{red}{\tiny ($\downarrow$  #1)}}
\newcommand{\same}[1]{\textcolor{blue}{\tiny ($-$  #1)}}

\definecolor{Gray}{gray}{0.90}

\definecolor{baselinecolor}{gray}{.9}
\newcommand{\baseline}[1]{\cellcolor{baselinecolor}{#1}}

\newcolumntype{x}[1]{>{\centering\arraybackslash}p{#1pt}}
\newcolumntype{y}[1]{>{\raggedright\arraybackslash}p{#1pt}}
\newcolumntype{z}[1]{>{\raggedleft\arraybackslash}p{#1pt}}

\newlength\savewidth\newcommand\shline{\noalign{\global\savewidth\arrayrulewidth
		\global\arrayrulewidth 1pt}\hline\noalign{\global\arrayrulewidth\savewidth}}
\newcommand{\tablestyle}[2]{\setlength{\tabcolsep}{#1}\renewcommand{\arraystretch}{#2}\centering\footnotesize}

\definecolor{iccvblue}{rgb}{0.21,0.49,0.74}
\usepackage[pagebackref,breaklinks,colorlinks,allcolors=iccvblue]{hyperref}

\def\paperID{10354} 
\def\confName{ICCV}
\def\confYear{2025}

\title{Reinforcement Learning meets Masked Video Modeling : Trajectory-Guided \\ Adaptive Token Selection}

\author{Ayush K. Rai$^{*,1}$, Kyle Min$^{*,2}$ \\ Tarun Krishna$^{1}$, Feiyan Hu$^{1}$, Alan F. Smeaton$^{1}$, Noel E. O'Connor$^{1}$\\
$^1$Insight Research Ireland Centre for Data Analytics, Dublin City University $^2$Intel Labs, USA \\
{\tt\small ayush.rai3@mail.dcu.ie    kyle.min@intel.com}
}

\begin{document}
\maketitle

\def\thefootnote{*}\footnotetext{Equal contribution}

\begin{abstract}
Masked video modeling~(MVM) has emerged as a highly effective pre-training strategy for visual foundation models, whereby the model reconstructs masked spatiotemporal tokens using information from visible tokens. However, a key challenge in such approaches lies in selecting an appropriate masking strategy. Previous studies have explored predefined masking techniques, including random and tube-based masking, as well as approaches that leverage key motion priors, optical flow and semantic cues from externally pre-trained models. In this work, we introduce a novel and generalizable \textbf{T}rajectory-Aware \textbf{A}daptive \textbf{T}oken \textbf{S}ampler (TATS), which models the motion dynamics of tokens and can be seamlessly integrated into the masked autoencoder (MAE) framework to select motion-centric tokens in videos. Additionally, we propose a unified training strategy that enables joint optimization of both MAE and TATS from scratch using Proximal Policy Optimization (PPO). We show that our model allows for aggressive masking without compromising performance on the downstream task of action recognition while also ensuring that the pre-training remains memory efficient. Extensive experiments of the proposed approach across four benchmarks, including Something-Something v2, Kinetics-400, UCF101, and HMDB51, demonstrate the effectiveness, transferability, generalization, and efficiency of our work compared to other state-of-the-art methods.


\end{abstract}

\begin{figure*}[t]
    \centering
    \includegraphics[width=1.0\textwidth]{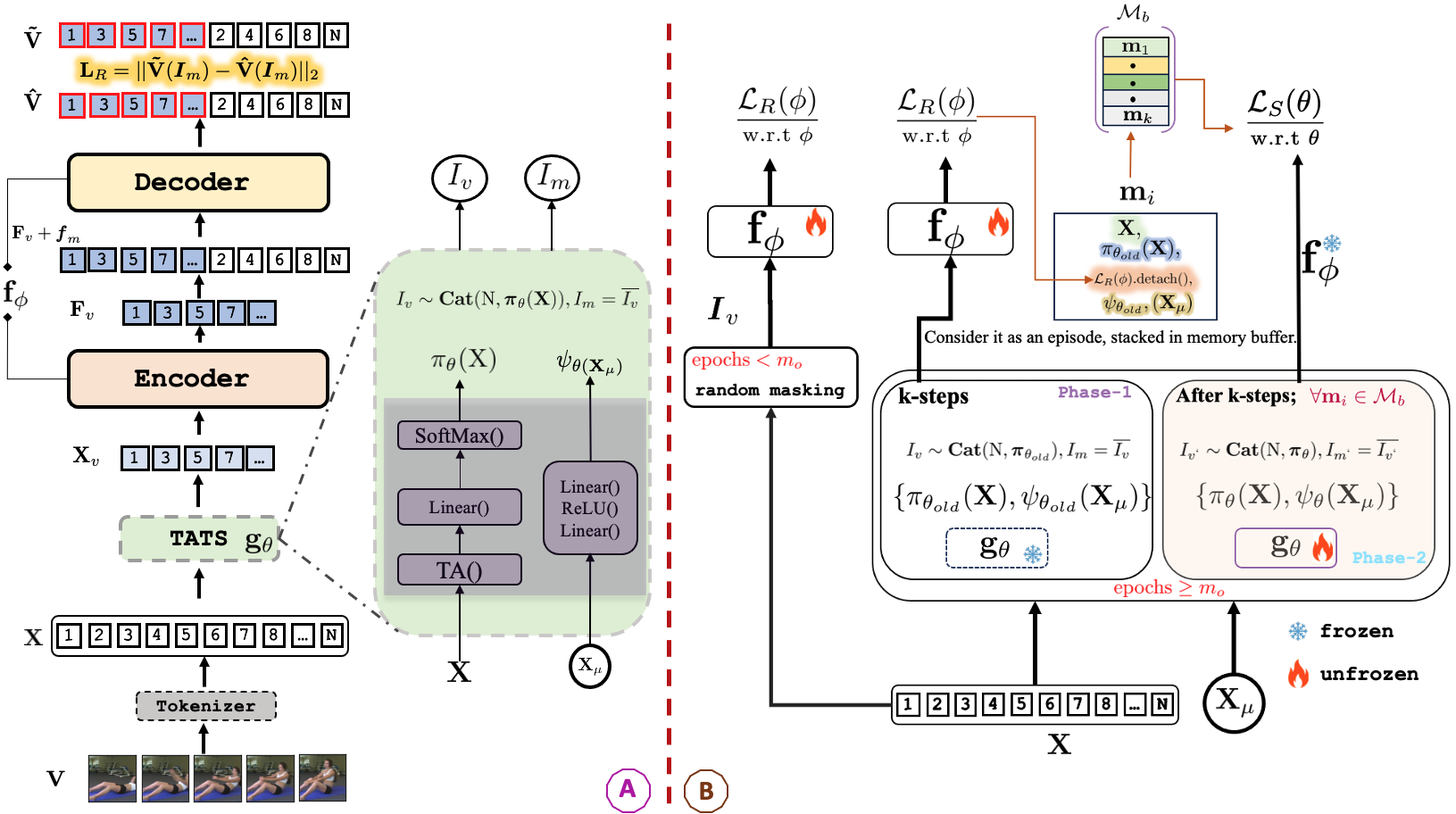}
    \small
    \caption{\textcolor{purple}{A} depicts our overall architecture with MAE ($f_{\phi}$) and TATS  ($g_\theta$). \textcolor{brown}{B} illustrates the joint training (Algorithm~\ref{ppo_training_recipe}) of $f_\phi$ and $g_\theta$ using PPO. Until epoch $m_o$, standard random space-time masking is applied. Afterward, every $k$ steps, Phase 1 ($g_\theta$ frozen,$f_\phi$ unfrozen) stores old state of $g_\theta$ in memory buffer $\mathcal{M}_b$ as episodes, followed by Phase 2 ($g_\theta$ unfrozen,$f_\phi$ frozen), where $g_\theta$ is optimized via $\mathcal{L}_{s}(\theta)$. The optimization process then alternates between Phase 1 and Phase 2.
} 
    \label{fig:main_arch}
\end{figure*}

\section{Introduction}
\label{sec:intro}

Self-supervised video representation learning has recently emerged as a prominent area of research due to the generalization capabilities of the learned embeddings. Such representations can be applied to several downstream tasks such as action recognition \cite{wang2022bevt,han2020self}, object detection \cite{devi2023}, and segmentation \cite{Aydemir2023NeurIPS} in videos. Due to the scarcity of labeled data, a standard approach in self-supervised learning (SSL) methods for video understanding involves defining a pretext task. A pretext task can be interpreted as a self-supervised pseudo-objective for pre-training a model. Intuitively, if a model learns to solve a complex task that requires a high-level understanding of its input, then the features learned as a result should generalize well to other tasks.

Inspired by BERT \cite{kenton2019bert} in language modeling, masked modeling in the form of masked autoencoders (MAE) was adopted for images \cite{wei2022masked,he2022masked,li2022semmae} and for videos \cite{feichtenhofer2022masked,tong2022videomae,wang2023videomae} as a self-supervised pretext task. Masked modeling involves masking a large fraction (between 75-95\%) of the input data and then learning to reconstruct or predict the removed content based on the visible information. Although this concept is simple, it has been shown to improve performance \cite{feichtenhofer2022masked, he2022masked}, generalization \cite{feichtenhofer2022masked, he2022masked}, data efficiency \cite{tong2022videomae}, memory efficiency~\cite{feichtenhofer2022masked, bandara2023adamae}, scalability~\cite{wang2023videomae, he2022masked}, robustness~\cite{hendrycks2019} and to reduce overfitting \cite{Girdhar_2023_CVPR} on downstream tasks.


Several studies have explored different formulations of MAE, focusing on masking portions of input, features, augmenting the masked modeling objective \cite{baobeit, dong2023peco, li2022semmae, xie2022simmim, zhou2022image, xie2023masked, wei2022masked}. However, less emphasis has been given to adaptive masking mechanisms that adaptively select space-time patches based on the input. The masking mechanism forms a crucial component of the family of MAE methods, as it is responsible for selecting which information is to be exploited by the encoder and predicted by the decoder.


\cite{he2022masked,xie2022simmim} explored random masking approaches for image patches, blocks, and grids. Though such approaches have shown promise and performance gains on downstream tasks, there still exists a research gap in terms of the masking mechanism being able to adapt to the input. With fixed masking mechanisms, MAEs are unable to exploit the expressivity of transformer-based encoders. In this direction, other contemporary works have investigated different masking strategies for images such as semantically guided masking~\cite{li2022semmae}, uniform sampling for pyramid-based vision transformer (ViT)~\cite{li2022uniform} and utilizing mask generators based on object priors ~\cite{chen2023improving} and learning easy-to-hard masking through curriculum learning~\cite{madan2024cl}. 

The challenging aspect of MVM is the extra time dimension and high spatiotemporal inductive biases from adjacent frames carrying highly redundant information. This introduces potential information leakage as masked space-time patches can be trivially inferred from spatiotemporal neighborhoods, enabling learning of the shortcuts and less generalizable representations during pre-training. Hence, a substantial amount of compute and memory resources are inefficiently utilized in the prediction of uninformative tokens. On the contrary, such a high level of redundancy can be exploited to aggressively mask space-time tokens with a high mask ratio without compromising the prediction quality of the masked space-time patches. 

Several approaches in MVM have utilized frame, tube, and patch-based masking~\cite{feichtenhofer2022masked, tong2022videomae, wang2023videomae}, and there is no single universal masking strategy that works for all datasets. VideoMAE~\cite{tong2022videomae} achieves the best action classification results on Something-Something v2 (SSv2)~\cite{goyal2017something} with random tube masking while STMAE~\cite{feichtenhofer2022masked} achieves its best performance on Kinetics-400 (K400)~\cite{kay2017kinetics} with random space-time patch masking. An explanation for this observation is that not all space-time tokens carry meaningful information, and a fixed masking strategy steers the model's optimization towards a particular task. Thus, it is crucial to incorporate adaptive computation in MAEs to dynamically select informative tokens based on the given input and the mask ratio. Previous works such as MGMAE \cite{huang2023mgmae} proposed motion-guided masking by extracting the optical flow from pre-trained models, and AdaMAE \cite{bandara2023adamae} introduced a token sampler module to select high-activity regions using REINFORCE \cite{Williams1992SimpleSG}. In order to exploit unequal information density among patches, we introduce \textit{TATS} module that learns a video-specific masking strategy from scratch to select space-time patches based on their spatio-temporal motion and trajectory information using Trajectory Attention (TA) \cite{patrick2021keeping}. \textit{TATS} does not rely on any computationally expensive dense optical flow features or semantic cues obtained from external pretrained models like  RAFT~\cite{teed2020raft}, CLIP~\cite{radford2021learning}, or DINOv2~\cite{oquab2023dinov2}. 

\textit{TATS} can be interpreted as a policy agent that models a categorical distribution over the set of input space-time tokens by leveraging their trajectory information and then samples the most relevant tokens based on a predefined mask ratio. However, since training MAE in conjunction with \textit{TATS} is unstable due to the non-differentiability of the sampling operation, we additionally propose a unified training recipe to train MAE and TATS modules simultaneously using PPO~\cite{schulman2017proximal} method used in reinforcement learning (RL). Our goal is to incorporate adaptivity into MAEs while preserving their representation quality in terms of generalization and ensuring that the pre-training process remains memory efficient.  Overall, our main contributions are:
\begin{itemize}
    \item We propose a novel and generalizable TATS module that learns to adaptively sample motion-centric tokens for MAE pre-training by modeling their motion trajectories in videos. TATS can be seamlessly integrated into the MAE framework and does not rely on auxiliary modalities like optical flow~(RAFT~\cite{teed2020raft}) or external pre-trained models (DINOv2~\cite{oquab2023dinov2}, CLIP~\cite{radford2021learning}) for motion or semantic cues.
    \item 
    Additionally, we introduce a unified training recipe (Algorithm~\ref{ppo_training_recipe}) that facilitates the joint optimization of both MAE and TATS from scratch using PPO~\cite{schulman2017proximal} to ensure stable convergence during pre-training even with aggressive masking.
    \item   Finally, we conduct a comprehensive evaluation on four benchmark datasets (K400, SSv2, UCF101, HMDB51) for action recognition to demonstrate the effectiveness, generalization, transferability, and efficiency of our work compared to the state-of-the-art methods (Tables~\ref{finetuning_results},\ref{transfer_results},\ref{tab:ablation_table}).
\end{itemize}

\section{Related Work}
\label{sec:related_work}

\subsection{SSL for video representation learning. }


SSL has emerged as a promising alternative to the supervised paradigm for pre-training deep models, enabling training on large-scale datasets with enhanced generalization while eliminating the need for labeled annotations. SSL in video primarily focuses on leveraging the temporal dimension for designing tasks such as temporal ordering \cite{fernando2017self, lee2017unsupervised, misra2016shuffle, wei2018learning, wang2019self}, future prediction \cite{vondrick2016anticipating, mathieu2016deep, lotter2017deep, vondrick2018tracking, diba2019dynamonet}, spatiotemporal contrast \cite{feichtenhofer2021large, han2019video, qian2021spatiotemporal, sun2019learning}, temporal coherence \cite{goroshin2015unsupervised,wiskott2002slow} and object motion \cite{agrawal2015learning, pathak2017learning, wang2015unsupervised, wang2019learning}.



\subsection{Masked Modeling. }
Masked Language Modeling has been universally adopted in natural language understanding, leading to groundbreaking works such as BERT~\cite{kenton2019bert}. Several researchers have adopted masked prediction for images/videos through Masked Image Modeling (MIM)/MVM, respectively. MIM/MVM can be interpreted as a generalized Denoising Autoencoder~\cite{denoising_AE} where the masking can be attributed to noise addition.

Generative Pre-training from pixels~\cite{pmlr-v119-chen20s} introduced the task of masked pixel prediction. However, pixel-level prediction demands high computational costs for pre-training and results in inferior performance compared to ConvNets. The notion of dividing an image into visual tokens through patches, as introduced in the ViT~\cite{dosovitskiy2020vit}, enabled the adoption of BERT-style pre-training for visual tokens. BeiT \cite{baobeit} and PeCo\cite{dong2023peco} are built upon using an offline tokenizer to learn discrete codebooks using VQ-VAE \cite{van2017neural} with the goal of reconstructing the original image from randomly masked discrete tokens. iBOT~\cite{zhou2022image} and DALL-E~\cite{ramesh2021zero} proposed an online tokenizer based on teacher networks trained via self-distillation.
Maskfeat \cite{wei2022masked} introduced reconstruction of Histogram-of-Oriented-Gradients features for masked-out regions. MAE \cite{he2022masked} and SimMIM~\cite{xie2022simmim} claimed that directly reconstructing the RGB pixel values performs equivalent to codebook-based methods. \cite{zhang2022mask} proposed a theoretical understanding of the masking mechanism.

 


\noindent MVM techniques have been employed for video pre-training by masking random space-time patches as in STMAE~\cite{feichtenhofer2022masked}, or by utilizing tube masking with a high masking ratio, as in VideoMAE~\cite{tong2022videomae, wang2023videomae}. Other MVM approaches include BEVT~\cite{wang2022bevt}, Masked Video Distillation~\cite{wang2023masked}. Our method is specifically designed for videos but can be integrated into the MAE framework while maintaining the original reconstruction target and MAE architecture without any modifications.

\subsection{Masking Strategies in MIM/MVM. }

Many studies have demonstrated that the performance of MAEs and their variants on downstream tasks relies heavily on the choice of masking strategy. 
SemMAE \cite{li2022semmae} harnesses iBOT~\cite{zhou2022image} for semantic segmentation and generates semantically aware masks for pre-training. ADIOS \cite{shi2022adversarial} introduces a method to jointly learn a masking function and an image encoder through an adversarial objective. AutoMAE \cite{chen2023improving} avails an adversarially-trained mask generator based on Gumbel-softmax~\cite{jang2016categorical} for MIM. CL-MAE \cite{madan2024cl} uses curriculum learning to generate adaptive masks based on the desired level of complexity (i.e. easy to hard masks). Cluster Masking \cite{wei2024efficient} learns to apply random masking to clusters of image patches, while R-MAE~\cite{nguyen2024rmae} focuses on masking pixels within a specified region. \cite{li2022uniform} proposed a uniform masking strategy that enables MAE pre-training for Pyramid-based ViTs with locality. AttnMask~\cite{kakogeorgiou2022what} proposed a distillation-based MIM 
where masking of the student network is guided by attention maps generated by the teacher network. \cite{xie2023masked} introduced a method to mask the frequency domain representation of the images using low/high pass filters while \cite{feng2023evolved} constructs a patch association graph using attention maps and addresses the unlabeled part partition problem as a graph cut problem using the Expectation-Maximization algorithm~\cite{banerjee2005clustering} to obtain semantics-based masks.

The masking strategy is a core design choice in MVM, which significantly impacts the information that the network learns during pre-training. MGMAE~\cite{huang2023mgmae} and MGM~\cite{fan2023motion} introduced motion-guided masking by exploiting a pre-trained lightweight optical flow estimator RAFT~\cite{teed2020raft} and motion vectors stored in the H.264 codec to select space-time patches with rich motion information. EVEREST~\cite{hwang2024everest} proposed redundancy robust token selection and an information-intensive frame selection mechanism for pre-training and fine-tuning. MME \cite{sun2023masked} modifies the pre-training objective from the reconstruction of the appearance content to the reconstruction of the motion trajectory. AdaMAE \cite{bandara2023adamae}, the work most closely related to ours, proposed an end-to-end trainable token sampling module that learns to sample space-time patches from high-activity regions using REINFORCE \cite{Williams1992SimpleSG}. Our approach draws inspiration from AdaMAE~\cite{bandara2023adamae}, however our \textit{TATS} module selects space-time tokens based on their motion trajectories in videos. Additionally, we propose a novel training recipe that jointly optimizes MAE and \textit{TATS} from scratch using PPO, ensuring stable convergence during pre-training, even with aggressive masking.

\section{Method}
\label{sec:method}

\subsection{Overview of MVM}


Here we briefly describe important components of a standard MVM method.

\noindent \textbf{Tokenizer.}~Consider an input video $V$ of size $T \times C \times H \times W$, where $T$ represents the number of frames, $C$ denotes the input channels and $H$, $W$ is the height and width of a frame. A \textit{Tokenizer} composed of 3D convolutional layer with kernel $K$ of size $( t,C,h,w)$, stride $S$ of size $(t,h,w)$ and $d$ output channels is availed to tokenize $V$ into $N$ number of tokens with dimension $d$ indicated as $\mathbf{X}$, where $N = \frac{T}{t} \times \frac{H}{h} \times \frac{W}{w}$. Positional information is further embedded into the tokens using a fixed 3D periodic positional encoding scheme outlined in \cite{vaswani2017attention}.




\noindent \textbf{Token Sampler.} Based on a specific masking mechanism (tube~\cite{tong2022videomae}, adaptive~\cite{bandara2023adamae}, random space-time~\cite{feichtenhofer2022masked}), a set of visible token indices $\boldsymbol{I}_v$ are sampled from $\boldsymbol{X}$ for a given mask ratio $\rho \in (0,1)$ while the remaining indices correspond to the masked Indices $\boldsymbol{I}_m$. The choice of masking mechanism is a pivotal design choice of MVM techniques. 

\noindent \textbf{Encoder-Decoder.} The design of encoder-decoder is usually a variant of VideoMAE~\cite{tong2022videomae}. The encoded representation $\mathbf{F}_v$ is learned by feeding the sampled visible tokens $\mathbf{X}_v$ into a vision transformer-based encoder. The learned representations for visible tokens  $\mathbf{F}_v$ are concatenated with a fixed learnable representation $f_m$ corresponding to the masked tokens. Subsequently, the positional information is added for both representations in the same order. These combined tokens are then passed through a transformer-based decoder to estimate predictions $\mathbf{\hat{V}}$. 

\noindent The entire network is trained by reconstruction loss computed between ground-truth and predicted values for the masked tokens.


\subsection{Trajectory-Aware Adaptive Token Sampler}
We propose \textit{TATS} module ($g_\theta$) that can be easily integrated into the family of MAE ($f_\phi$) architectures, can be trained from scratch and learns to sample motion-centric tokens without the use of any external pre-trained models to compute optical flow such as RAFT~\cite{teed2020raft} in MGMAE~\cite{huang2023mgmae}, motion vector in MGM~\cite{fan2023motion} or having motion-specific pre-training objective in MME~\cite{sun2023masked}. In particular, we avail of TA~\cite{patrick2021keeping}, which captures motion dynamics in a video by learning a probabilistic path of a token between frames. By exclusively sampling motion-centric tokens, \textit{TATS} facilitates the encoder to learn more generic and expressive representations, which is crucial for downstream tasks such as action recognition. The computational overhead of \textit{TATS} is minimal compared to MAE.

\noindent \textbf{Trajectory Attention.}~In the \textit{TATS} module, we apply TA \cite{patrick2021keeping} across space-time tokens, where a trajectory represents the probabilistic path of a token in a video sequence determined by the motion between a pair of frames. A set of query-key-value vectors $\textbf{q}_{st}, \textbf{k}_{st}, \textbf{v}_{st} \in \mathbb{R}^{d}$ are computed through linear projections ($W$) for a given space-time token \(\mathbf{x}_{st}\) ($\mathbf{x}_{st} \in \mathbf{X}$) corresponding to space-time location $st$ (`reference point') in a video. For \(\mathbf{q}_{st}\), a set of trajectory tokens $\tilde{\mathbf{y}}_{stt'} \in \mathbb{R}^{d}$ are  computed, encapsulating spatially pooled information weighted by the trajectory probability. These trajectory tokens, extend throughout the video sequence and can be represented independently at different time steps $t'$.



\begin{equation}
\tilde{\mathbf{y}}_{stt'}
=
\sum_{s'}
\mathbf{v}_{s't'} \cdot
\frac
{\exp\langle \mathbf{q}_{st}, \mathbf{k}_{s't'}\rangle}
{\sum_{\bar s}\exp \langle \mathbf{q}_{st}, \mathbf{k}_{\bar st'}\rangle}.
\end{equation}






\noindent Here, $\exp$ denotes the exponential function, $\langle.,.\rangle$ represents dot product, and $\cdot$ indicates element-wise multiplication. Next, trajectories are pooled across time to capture intra-frame relationships. The trajectory tokens \(\tilde{\mathbf{y}}_{stt'}\) are mapped to a new set of query, key, and value representations, denoted as \(\tilde{\mathbf{q}}_{st}, \tilde{\mathbf{k}}_{stt'}, \tilde{\mathbf{v}}_{stt'}\), using the projection matrix \(\tilde{W}\). Now \(\tilde{\mathbf{q}}_{st}\) becomes the updated  reference query for reference point \(st\).  $\tilde{\mathbf{q}}_{st}$ is then utilized to aggregate information across the temporal dimension using 1D attention given by:

\begin{equation}
    \mathbf{y}_{st}
    =
    \sum_{t'}
    \tilde{\mathbf{v}}_{stt'} \cdot
    \frac
    {\exp\langle \tilde{\mathbf{q}}_{st}, \tilde{\mathbf{k}}_{stt'}\rangle}
    {\sum_{\bar t}\exp \langle \tilde{\mathbf{q}}_{st}, \tilde{\mathbf{k}}_{st\bar t}\rangle}.
\end{equation}



\noindent The trajectory information is encoded in $\mathbf{y}_{st}$. In practice, we employ an approximation of TA (Orthoformer~\cite{patrick2021keeping}), which has linear complexity in space-time.

\noindent The \textbf{TATS module} ($g_\theta$) has two branches, one of which processes the input tokens $\text{X}$ by passing them through a block of TA followed by a \textit{Linear} layer, and a \textit{Softmax} activation to compute the probability scores $\pi_{\theta}(X) \in \mathbb{R}^{N}$ for all tokens.

\begin{equation}
    \boldsymbol{Z} = \text{TA} (\boldsymbol{X}) \text{;  }  \text{Z} \in \mathbb{R}^{N \times d} \\
\end{equation}
\begin{equation}
    \pi_{\theta}(X) = \texttt{Softmax}(\texttt{Linear}(\boldsymbol{Z})) \in \mathbb{R}^{N}
\end{equation}


\noindent Following AdaMAE~\cite{bandara2023adamae}, these probability scores are utilized to define an $N$-dimensional categorical distribution over $\pi_{\theta}(X)$, from which visible token indices are sampled without replacement i.e. $\boldsymbol{I}_{v} \sim \text{Categorical}(N, \pi_{\theta}(X))$. The masked token indices are the complement of visible token indices and are given by $\boldsymbol{I}_m = \overline{\boldsymbol{I}_v}$. The number of sampled visible tokens $\boldsymbol{N}_{v} = N \times (1 - \rho)$ and $\rho \in (0,1)$ is the predefined mask ratio. This branch can be interpreted as the actor-network (or policy network), which outputs the probability of relevance for every token. In other words, this output probability can be perceived as policy $\pi_\theta(\text{X})$ representing the likelihood of a token being selected given its token representation $\boldsymbol{X}$. The second branch processes the mean representation of all the tokens ($X_\mu$) and passes it through a feed-forward network consisting of two linear layers with a ReLU activation $\text{Linear}(1568) \rightarrow \text{ReLU}(\text{Linear}(784)) \rightarrow 1$. This can be interpreted as the value network, which learns to predict the expected reward for the current input tokens $\boldsymbol{X}$, given a mean state $X_\mu$. We denote the output of the value network as $\psi_{\theta}(X_\mu)$. This value is used for computing the advantage $\boldsymbol{A}(X,I_m)$ as detailed in the optimization section. Overall, the computation of \textit{TATS} ($g_\theta$) can be represented as:
\begin{equation}
    \label{tats_equation}
    \pi_\theta(X), \mathbf{\psi}_\theta(X_\mu) = g_\theta(X)
\end{equation}
The complete architecture is shown in Fig~\ref{fig:main_arch}. 

\subsection{Optimization}




\textit{TATS} can be conceptualized as an agent interacting with its environment, represented by the MAE, with the objective of learning an optimal masking strategy that removes redundant tokens while selecting only the most informative and motion-centric ones for encoding, given a mask ratio $\rho$. The environment provides feedback to \textit{TATS} through a reward, which corresponds to the reconstruction error $\mathcal{L}_{R}$ \cite{bandara2023adamae}. 

\noindent The intuition for this reward is that tokens with low reconstruction errors are easier to reconstruct and thus contain redundant information, whereas motion-centric tokens, which are more challenging to reconstruct, exhibit higher reconstruction error. Consequently, \textit{TATS} must be optimized to prioritize the selection of these motion-centric tokens or tokens with higher reconstruction error. Our optimization strategy is loosely inspired from application of RL~\cite{goldberg2023rlhf} in the context of aligning large language model (LLM) outputs with human preferences. A major challenge in this formulation is the simultaneous training of both the agent (\textit{TATS}) and the reward model (\textit{MAE}), differing from conventional LLM approaches where the reward model is typically pre-trained separately based on human-labeled data. The joint optimization process incorporates two distinct losses, i.e. reconstruction loss and the sampling loss, as outlined below.


\noindent \textbf{Reconstruction Loss}: To optimize the MAE (characterized by $f_\phi$), we compute the mean squared error loss $\mathcal{L}_R$ between the predicted and the normalized ground-truth RGB values of the masked tokens as shown in the following equation:  

\begin{equation}
    \label{reconstruction_loss}
    \mathcal{L}_{R} (\phi) = \frac{1}{N - N_v} \sum_{i \in \boldsymbol{I}_m} || \mathbf{\tilde{V}}_i - \mathbf{\hat{V}}_i ||_2
\end{equation}

\noindent where $\mathbf{\hat{V}}$ denotes the predicted tokens from the decoder, $\mathbf{\tilde{V}}$ represents the patch normalized ground-truth RGB values corresponding to the masked tokens.

\noindent \textbf{Sampling Loss.} \textit{TATS} ($g_\theta$) is optimized using the sampling loss $\mathcal{L}_S(\theta)$ based on PPO~\cite{schulman2017proximal}. To jointly train $f_\phi$ and $g_\theta$ from scratch, we propose a unified training approach that alternates between optimizing  $f_\phi$ and $g_\theta$. Initially, our objective is to train $f_\phi$ up to epoch $m_o$ using random space-time masking, minimizing the reconstruction loss $\mathcal{L}_R$. This ensures that the MAE learns the task of reconstructing masked tokens, as the reconstruction error would be used as a reward for sampling the most challenging space-time tokens.




\noindent Since $g_\theta$ is trained using PPO, which requires episodes recorded from a previous state of $g_\theta$. To facilitate this during Phase~1~(after $m_o$ epochs), for every $k$ steps, $g_\theta$ is kept frozen while $I_v \sim \text{Categorical}(N,\pi_{\theta_\text{old}}(X))$ and $f_\phi$ is optimized based on $\mathcal{L}_{R}{(\phi)}$. Simultaneously, the memory buffer $\mathcal{M}_b$ is updated with recorded episodes in the form of $\{X, \pi_{\theta_\text{old}}(I_m \vert X), \mathcal{L}_R(\phi)\text{.detach}, \mathbf{\psi}_{\theta_\text{old}}(X_\mu)\}$. Here, $X$ represents the tokens, $\pi_{\theta_\text{old}}(I_m \vert X)$ denotes the probability of sampling the masked indices, $\mathcal{L}_R(\phi)$ corresponds to the reconstruction error from $f_\phi$, and $\mathbf{\psi}_{\theta_\text{old}}(X_\mu)$ represents the output of the value network. Using recorded rewards and value estimates, the advantage is computed as $A_{\theta_{\text{old}}} (\text{X},I_m) = \mathcal{L}_{R} (\phi) - \mathbf{\psi}_{\theta_{\text{old}}}(\text{X}_\mu)$.

\noindent In Phase~2, $f_\phi$ is frozen while $g_\theta$ is unfrozen. Recorded episodes are then sampled from $\mathcal{M}_b$, and the current state of $g_\theta$ is used for computing  $\pi_\theta(X), \mathbf{\psi}_\theta(X_\mu) = g_\theta(X)$ and $I_{v'} \sim \text{Categorical}(N,\pi_\theta(X))$. Notably, $\mathcal{L}_{R} (\phi)$ is detached from the computation graph to prevent gradient propagation in MAE during this step. The overall PPO objective used for training $g_\theta$ is defined by the following equation.


\begin{equation}
\begin{aligned}
    \mathbf{\mathit{J}}^{PPO} (\theta) = \mathbb{E} \Big[ &c_1 \mathbf{\mathit{J}}^{\text{CLIP}} (\theta) 
    - c_2 \left( \mathbf{\psi}_\theta(\text{X}_\mu) - \mathcal{L}_{R} (\phi) \right)^2 \\
    &+ c_3 \mathbf{H} (\text{X}, \pi_\theta)(.) \Big]
\end{aligned}
\label{ppo_objective}
\end{equation}
\begin{equation}
\begin{aligned}
     \mathbf{\mathit{J}}^{\text{CLIP}} (\theta) = \mathbb{E} \Big[ \text{min} &\left( r(\theta) A_{\theta_{\text{old}}} (\text{X},I_m), \right. \\
     &\left. \text{clip} \left( r(\theta), 1 - \epsilon, 1 + \epsilon \right) A_{\theta_{\text{old}}} (\text{X},I_m) \right) \Big]
\end{aligned}
\label{ppo_clip_objective}
\end{equation}

where $r(\theta) = \frac{\pi_\theta(I_{m'} \vert \text{X})}{\pi_{\theta_\text{old}}(I_m \vert \text{X})}$ represents the importance sampling ratio, and $\epsilon = 0.2$ is the clipping (clip) threshold. The term $(\mathbf{\psi}_{\theta}(\text{X}_\mu) - \mathcal{L}_{R} (\phi))^2$
serves as the objective for training the value network, representing the error in value estimation. $\mathbf{H} (\text{X}, \pi_\theta)(.)$ denotes the entropy term associated with the tokens $X$ and policy $\pi_\theta$, promoting sufficient exploration. The coefficients $c_1, c_2, c_3$ balance $\mathbf{\mathit{J}}^{\text{CLIP}}(\theta)$ (policy loss), value loss, and entropy term, respectively, in the overall PPO objective. After completing Phase 2, $f_\phi$ is unfrozen, $\mathcal{M}_b$ is reset, and the algorithm transitions back to Phase 1. This alternating process continues, switching between Phase 1 and Phase 2 iteratively throughout training. Since we want to minimize the sampling loss hence $\mathcal{L}_S(\theta) = -\mathbf{\mathit{J}}^{PPO} (\theta)$. AdaMAE \cite{bandara2023adamae} utilizes REINFORCE~\cite{Williams1992SimpleSG} which has high variance, however using PPO \cite{schulman2017proximal} improves stability as it uses a clipped objective $\mathbf{\mathit{J}}^{\text{CLIP}}(\theta)$ preventing it from making large updates, therefore balancing exploration and exploitation. Our training recipe is illustrated in Algorithm \ref{ppo_training_recipe}.

\begin{algorithm} \small
\caption{Unified Training Recipe for joint optimization of MAE and TATS.}
\begin{algorithmic}[1]

\Require Video $V$, MAE network $f_\phi$, \textit{TATS} module $g_\theta$, mask ratio $\rho$, memory buffer $\mathcal{M}_b$, epochs $E$, Train only MAE epochs $m_o$, \textit{TATS} update interval $k$, Total number of tokens $N$.
\State Initialize MAE $f_\phi$ and \textit{TATS} $g_\theta$.

\For{$e = 1$ to $E$}
    \For{$\text{step, batch in dataloader}$}
    \State tokenize $V$ into $X$ with indices $\{\boldsymbol{I}_1, \boldsymbol{I}_2, \dots, \boldsymbol{I}_N\}$.
        \If{$e \leq m_o$} \Comment{Random Space-Time Masking Phase}
            \State $I_v \sim$ \text{Random Distribution with } $\rho$ 
            \State optimize $\mathcal{L}_{R} = f_\phi(X_v)$ w.r.t. $\phi$.
        \Else \Comment{TATS Training Phase}
            \State freeze $g_\theta$, compute $\pi_{\theta_{\text{old}}}(X), \mathbf{\psi}_{\theta_\text{old}}(X_\mu)$ = $g_\theta(X)$.
            \State $I_v \sim \text{Categorical}(N,\pi_{\theta_\text{old}}(X)) \text{ with } \rho$ ; $I_m = \overline{I_v}$ 
            \State optimize $\mathcal{L}_{R}(\phi) = f_\phi(X_v)$ w.r.t. $\phi$. 
            \State episode = $\{X, \pi_{\theta_\text{old}}(I_m \vert X), \mathcal{L}_R (\phi)\text{.detach}, \mathbf{\psi}_{\theta_\text{old}}(X_\mu)\}$
            \State $\mathcal{M}_b$\text{.update(episode)}
            \If{$\text{step} \bmod k = 0$} \Comment{TATS Update}
                \State freeze $f_\phi$, unfreeze $g_\theta$.
                \For{episode in $\mathcal{M}_b$}
                    \State compute $\pi_\theta(X), \mathbf{\psi}_\theta(X_\mu)$ = $g_\theta(X)$
                    \State $I_{v'} \sim \text{Categorical}(N,\pi_{\theta}(X))$ ; $I_{m'} = \overline{I_{v'}}$
                    \State optimize $\mathcal{L}_S(\theta) = -\mathbf{\mathit{J}}^{PPO} (\theta)$ w.r.t. $\theta$.
                \EndFor
                \State unfreeze $f_\phi$.
                \State $\mathcal{M}_b$\text{.reset()}
            \EndIf
        \EndIf
    \EndFor
\EndFor

\end{algorithmic}
\label{ppo_training_recipe}
\end{algorithm}

\section{Experimental Setup}
\label{sec:experimental_setup}

\textbf{Datasets.}~We validate our method on four benchmarks:~SSv2~\cite{goyal2017something}, K400~\cite{kay2017kinetics}, UCF101~\cite{soomro2012ucf101} and HMDB51~\cite{kuehne2011hmdb}. 

\noindent \textbf{Implementation Details.} We employ the ViT-Base model ($\approx$87M parameters)~\cite{dosovitskiy2020vit} for our experiments. The input video has the dimension $16 \times 3 \times 224 \times 224$ while the patch size is $2 \times 3 \times 16 \times 16$ (tubelet length $= 2$), yielding a total of $1568$ tokens. Mask ratio $\rho$ takes the value $\{0.85, 0.90, 0.95\}$. Our experiments contain two types of settings:

\noindent \textbf{1.~Small Scale Pre-training.}~For K400 and SSV2, we construct a smaller training data subset by sampling approximately 15\% of the training set (equivalent to validation set size), while maintaining a class distribution consistent with the original dataset. Notably, the validation set remains unchanged from the original dataset. The standard train/validation sets for UCF101 and HMDB51 are used. The models have been pre-trained for $400$ epochs with a batch size of $32$ on $8$ Nvidia A100 GPUs.

\noindent \textbf{2.~Large Scale Pre-training}~is also conducted on full SSv2, however due to computational constraints, we only pretrain it for 400 epochs and $\rho=0.95$ on $8$ Nvidia A100 GPUs.



\noindent \textbf{Evaluation on action recognition.}~To assess the effectiveness of the pre-trained encoder, we conduct end-to-end \textit{fine-tuning} for the action recognition task over 100 epochs with the evaluation metric being top-$1$ and top-$5$ accuracy. Most of our experiments are conducted in a small-scale setting, while results for large-scale pre-training and fine-tuning are explicitly reported. 

Refer to the supplementary material (supp) for additional implementation details. Source code for this work has been released at \href{https://github.com/rayush7/rl_videomae}{\texttt{github.com/rayush7/rl\_videomae}}.

\begin{table}[t]
 \small
 \caption{\small Comparison of fine-tuning result of \colorbox{baselinecolor}{ Our} model against baselines (\cite{bandara2023adamae,tong2022videomae}) on action recognition task across benchmark datasets and different $\rho$ with top-1/top-5 accuracy as evaluation metric. (\up{} / \down{} : denotes increase/decrease in performance)}.
 \label{finetuning_results}
 \centering
 \resizebox{1.0\linewidth}{!}{
 \begin{tabular}{l|c|cc|cc|cc}
\hline
\hline 
 Dataset &  Mask Ratio & \multicolumn{2}{c|}{VideoMAE~\cite{tong2022videomae}} & \multicolumn{2}{c|}{AdaMAE~\cite{bandara2023adamae}} & \multicolumn{2}{c}{\textbf{Ours}} \\
   & $\rho$ & top-$1$ & top-$5$ & top-$1$ & top-$5$ & top-$1$ & top-$5$\\
   \hline 
   
   & 0.85 & 80.36 & 94.95 & 83.98 & 96.37 & \baseline{85.94} \up{}  & \baseline{96.98} \up{} \\
 UCF101  & 0.90 & 76.64 & 94.29 & 82.42 & 95.84 & \baseline{84.53} \up{} & \baseline{96.37} \up{} \\
   & 0.95 & 65.86 & 89.14 & 80.83 & 95.26 & \baseline{81.75} \up{} & \baseline{95.29} \up{} \\
   \hline 
   & 0.85 & 40.82 & 71.61 & 41.28 & 73.37 & \baseline{41.60} \up{} & \baseline{73.31} \down{} \\
 HMDB51  & 0.90 & 36.39 & 69.73 & 39.13 & 72.33 & \baseline{41.28} \up{} & \baseline{73.76} \up{} \\
   & 0.95 & 33.98 & 65.36 & 37.70 & 70.38 & \baseline{38.67} \up{} & \baseline{72.01} \up{} \\
   \hline
   & 0.85 & 42.26 & 68.28 & 38.97 & 64.68 & \baseline{43.24} \up{} & \baseline{68.76} \up{} \\
 Kinetics-400  & 0.90 & 41.79 & 68.62 & 39.50 & 65.70 & \baseline{43.28} \up{} & \baseline{68.85} \up{} \\
   & 0.95 & 39.73 & 66.15 & 39.42 & 65.14 & \baseline{41.70} \up{} & \baseline{67.29} \up{}\\
   \hline
   & 0.85 & 37.63 & 66.47 & 37.92 & 66.63 & \baseline{39.96} \up{} & \baseline{68.10} \up{} \\
 SSv2  & 0.90 & 37.85 & 66.86 & 38.10 & 66.29 & \baseline{40.79} \up{} & \baseline{69.30} \up{} \\
   & 0.95 & 37.24 & 65.92 & 38.38 & 67.11 & \baseline{40.25} \up{} & \baseline{68.73} \up{} \\
    \hline
    \hline 
 \end{tabular}
 }
 \label{finetuning_results_table1}
 
 \end{table}

 \begin{figure*}[t]
    \centering
    \includegraphics[width=0.95\textwidth]{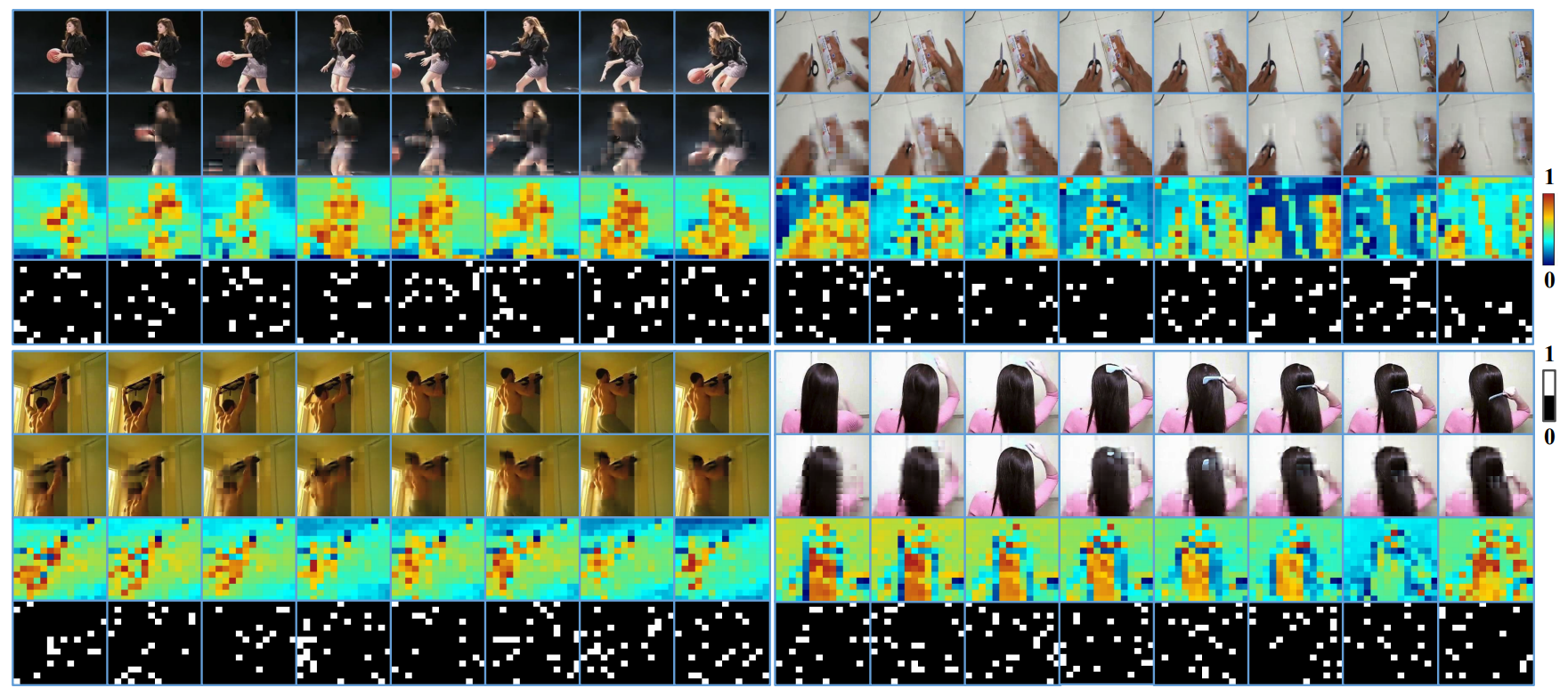}
    \caption{Visualization of adaptive masks learned by \textit{TATS} for $\rho$ = 0.95. The figure has four blocks: \textcolor{darkgreen}{top-left (K400)}, \textcolor{purple}{top-right (SSv2)}, \textcolor{brown}{bottom-left (UCF101)}, and \textcolor{blue}{bottom-right (HMDB51)}. In each block, the \underline{first row} shows video frames, the \underline{second} presents predictions/reconstructions, the \underline{third} depicts sampling probabilities for space-time tokens, and the \underline{fourth} displays the learned adaptive binary masks.}
    \label{fig:ours_mask_0.95}
\end{figure*}

\subsection{Results}

We perform extensive quantitative and qualitative studies of our approach on the given datasets and compare the performance against \cite{bandara2023adamae} and~\cite{tong2022videomae}~(baselines) respectively. For fair comparison with our method under \underline{small-scale pre-training} setup, these baselines were also pretrained (finetuned) for 400 (100) epochs on the same subset (K400/SSv2) using their public source code and default configuration. For \underline{large scale pre-training} results refer to the supp.

\noindent \textbf{Fine-tuning Results.}~ 
Table~\ref{finetuning_results} presents the top-1 and top-5 accuracy obtained after fine-tuning our method across different mask ratios, $\rho = \{0.85, 0.90, 0.95\}$. Our approach consistently surpasses~\cite{bandara2023adamae,tong2022videomae} across all benchmark datasets and mask ratios with the exception of top-$5$ accuracy on HMDB51 with $\rho=0.85$, which is marginally less than \cite{bandara2023adamae}. Notably, even under an aggressive masking ratio ($\rho = 0.95$), our model demonstrates superior performance compared to these baselines. These results highlight the effectiveness and generalization capability of the proposed \textit{TATS} module and the training strategy in terms of learning a better representation quality than learnt by \cite{bandara2023adamae,tong2022videomae}.


\noindent \textbf{Transferability.}~
Table~\ref{transfer_results} presents the transfer performance of our model on the action recognition task, pre-trained and fine-tuned across different datasets and mask ratio combinations. Our approach achieves better results than \cite{bandara2023adamae, tong2022videomae} across most settings, providing further insight into the strong transferability and generalization of our model.

\begin{figure}[t]
    \centering
    \includegraphics[width=0.5\textwidth]{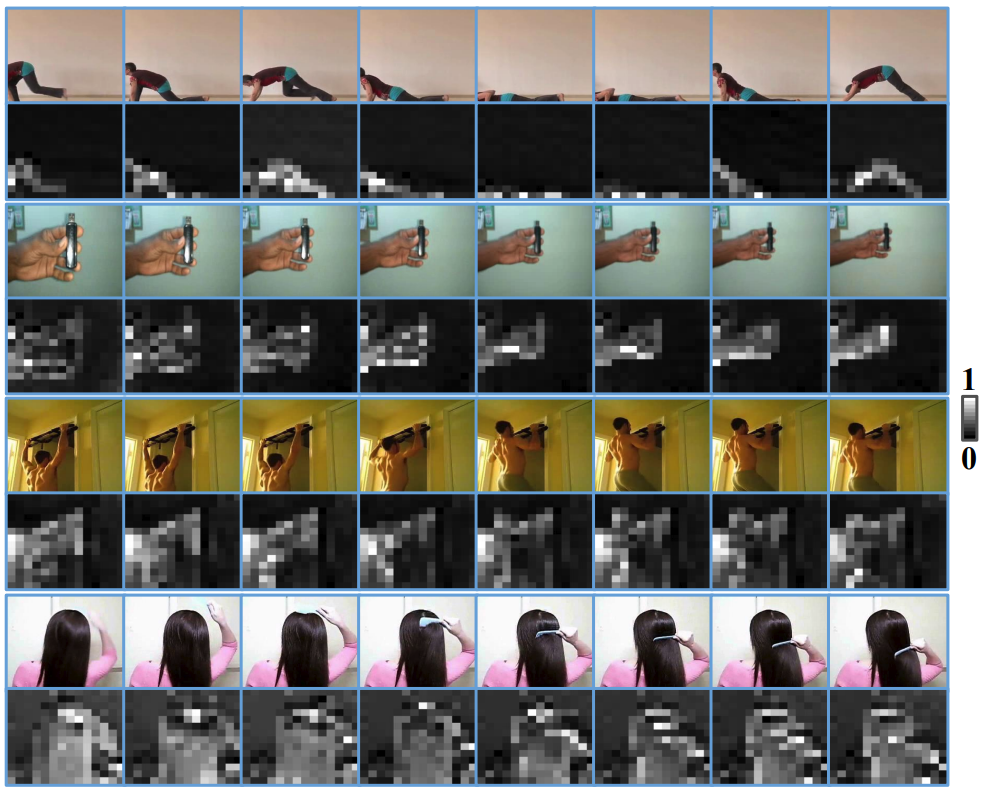}
    \caption{Visualization of the TA learnt by \textit{TATS}. The figure comprises four blocks : \textcolor{darkgreen}{K400}, \textcolor{purple}{SSv2}, \textcolor{brown}{UCF101}, and \textcolor{blue}{HMDB51} in top to bottom order. In each block, the \underline{first row} shows video frames, the \underline{second} depicts the trajectory attention on space-time tokens averaged across different heads.}
    \label{fig:traj_attn_fig}
\end{figure}

\noindent \textbf{Qualitative Assessment.}~We conduct a qualitative analysis by visualizing the learned adaptive binary masks learned by the \textit{TATS} module across the benchmark datasets and different $\rho$, as shown in Figure~\ref{fig:ours_mask_0.95}. We observe that \textit{TATS} learns to sample motion-centric tokens while also undergoing sufficient exploration enabling better generalization. Additionally, we visualize the learned TA across all space-time patches by averaging all heads, as depicted in Figure~\ref{fig:traj_attn_fig}. It is quite evident that our \textit{TATS} module accurately models motion trajectories of the space-time tokens as they evolve over time in the video, thereby enabling the sampling of motion-centric space-time patches. This also validates the formulation of the $\mathcal{L}_s$ and the training recipe to jointly train MAE and \textit{TATS}. Refer to supp for more mask visualizations.

\subsection{Ablation Studies}
We carry out an ablation study on UCF101 using models pre-trained with $\rho=0.95$ for 400 epochs and fine-tuned on the action recognition task for 100 epochs. The ablation results are illustrated in Table~\ref{tab:ablation_table}.

\noindent \textbf{1. Effect of Trajectory Attention.} 
In Table~\ref{tab:traj_effect}, we analyze the effect of integrating TA within the \textit{TATS} module compared to the Multi-Head Self-Attention
(MHA). Our findings indicate that TA achieves a top-1 accuracy of $81.75\%$ while utilizing $25.36$ GB of memory, outperforming MHA. This highlights the efficiency of TA in delivering superior performance with reduced memory consumption. Furthermore, our results also validate that TA effectively captures motion trajectories in a self-supervised manner, without relying on any motion-specific learning objective.

 \begin{table}[t]
 \small
 \caption{\small Comparison of transfer learning result of \colorbox{baselinecolor}{ Our} model against \cite{bandara2023adamae,tong2022videomae} on action recognition across benchmark datasets and different $\rho$ with top-1/top-5 accuracy as evaluation metric. (\up{} / \down{} / \same{} : denotes increased/decreased/equivalent performance)}
 \label{transfer_results}
 \centering
 \resizebox{1.0\linewidth}{!}{
 \begin{tabular}{l|c|cc|cc|cc}
\hline
\hline 
 Dataset &  Mask Ratio & \multicolumn{2}{c|}{VideoMAE~\cite{tong2022videomae}} & \multicolumn{2}{c|}{AdaMAE~\cite{bandara2023adamae}} & \multicolumn{2}{c}{\textbf{Ours}} \\
 From $\rightarrow$ To  & $\rho$ & top-$1$ & top-$5$ & top-$1$ & top-$5$ & top-$1$ & top-$5$ \\
   \hline 
   
   & 0.85 & 84.91 & 96.51 & 85.49 & 96.93 & \baseline{86.94} \up{} & \baseline{97.67} \up{}\\
 Kinetics-400 $\rightarrow$ UCF101  & 0.90 & 84.41 & 96.25 & 84.98 & 96.48 & \baseline{86.23} \up{} & \baseline{97.27} \up{} \\
   & 0.95 & 82.40 & 95.80 & 84.03 & 96.50 & \baseline{85.17} \up{} & \baseline{96.77} \up{}\\
   \hline 
   & 0.85 & 55.60 & 82.55 & 55.79 & 84.44 & \baseline{60.81} \up{} & \baseline{84.44} \same{} \\
 Kinetics-400 $\rightarrow$ HMDB51  & 0.90 & 56.71 & 83.07 & 56.45 & 82.49 & \baseline{60.42} \up{} & \baseline{83.59} \up{}\\
   & 0.95 & 53.26 & 79.75 & 54.10 & 81.25 & \baseline{58.14} \up{} & \baseline{82.62} \up{} \\
   \hline
    & 0.85 & 36.42 & 65.50 & 36.72 & 65.72 & \baseline{38.39} \up{} & \baseline{66.47} \up{} \\
 Kinetics-400 $\rightarrow$ SSv2  & 0.90 & 35.70 & 64.46 & 36.62 & 65.27 & \baseline{39.46} \up{} & \baseline{67.25} \up{} \\
   & 0.95 & 34.11 & 62.64 & 36.88 & 65.64 & \baseline{38.13} \up{} & \baseline{66.48} \up{} \\
   \hline 
   & 0.85 & 84.88 & 96.91 & 84.98 & 96.72 & \baseline{87.16} \up{} & \baseline{97.38} \up{} \\
 SSv2 $\rightarrow$ UCF101  & 0.90 & 83.88 & 96.75 & 84.64 & 97.06 & \baseline{86.81} \up{} & \baseline{97.51} \up{} \\
   & 0.95 & 82.53 & 95.90 & 84.38 & 96.21 & \baseline{85.14} \up{} & \baseline{97.11} \up{} \\
   \hline
   & 0.85 & 54.82 & 82.03 & 55.47 & 82.81 & \baseline{59.64} \up{} & \baseline{84.83} \up{}\\
 SSv2 $\rightarrow$ HMDB51  & 0.90 & 55.92 & 83.40 & 55.86 & 84.31 & \baseline{60.35} \up{} & \baseline{85.42} \up{} \\
   & 0.95 & 52.41 & 80.14 & 54.69 & 84.18 & \baseline{58.40} \up{} & \baseline{83.59} \down{}\\
    \hline
    \hline 
 \end{tabular}
 }
 \label{transfer_results_table1}
 \end{table}




\begin{table*}[t]
    \vspace{-.2em}
    \centering
    \subfloat[
    \textbf{Effect of Trajectory Attention}. Better performance is obtained with TA with marginally less memory usage.
    \label{tab:traj_effect}
    ]
    { 
    \begin{minipage}{0.34\linewidth}{\begin{center}
    \tablestyle{4pt}{1.05}
    \begin{tabular}{y{20}x{18}x{18}x{18}x{35}}
    \shline
    case & ratio & top-1 & top-5 & memory\\
    \shline
    MHA & 0.95 & 81.59 & 95.29 & 25.37 GB\\
    \baseline{TA} & \baseline{\textbf{0.95}} & \baseline{\textbf{81.75}} & \baseline{\textbf{95.42}} & \baseline{\textbf{25.36 GB}} \\
    \end{tabular}
    \end{center}}\end{minipage}
    }
    \hspace{0.5em}
    \subfloat[
    \textbf{Different decoder depth}. Our method performs best when \# of decoder blocks $=$ 4.
    \label{tab:decoder_depth}
    ]
    {
    \centering
    \begin{minipage}{0.24\linewidth}{\begin{center}
    \tablestyle{4pt}{1.05}
    \begin{tabular}{x{20}x{18}x{24}x{32}}
    \shline
    blocks & top-1 & top-5 & memory\\
    \shline
    1 &  81.46 & 95.07 & 16.54 GB \\ 
    2 &  80.83 & 95.02 & 19.48 GB \\
    \baseline{4} & \baseline{\textbf{81.75}} & \baseline{\textbf{95.29}} & \baseline{\textbf{25.36 GB}} \\
    8 & 79.10 & 94.68 & 37.13 GB \\
    \end{tabular}
    \end{center}}\end{minipage}
    }
    \hspace{2em}
    \subfloat[
    \textbf{Memory Usage}. Our method uses less memory (pretraining) than \cite{bandara2023adamae} while achieving significantly higher performance (finetuning) than \cite{tong2022videomae}.
    \label{tab:memory_usage}
    ]
    {
    \centering
    \begin{minipage}{0.23\linewidth}{\begin{center}
    \tablestyle{4pt}{1.05}
    \begin{tabular}{x{50}x{35}x{20}}
    \shline
    method & memory & top-1\\
    \shline
    VideoMAE~\cite{tong2022videomae} & 20.94 GB & 65.86\\
    AdaMAE~\cite{bandara2023adamae} &   26.17 GB & 80.83\\
    \baseline{\textbf{Ours}} & \baseline{\textbf{25.36 GB}} & \baseline{\textbf{81.75}} \\
    \end{tabular}
    \end{center}}\end{minipage}
    }
\\
    \centering
    \vspace{.5em}
    \subfloat[
    \textbf{Reconstruction Loss function}. The best result is obtained by optimizing MSE loss with local patch normalization.
    \label{tab:loss_target}
    ]
    {
    \begin{minipage}{0.36\linewidth}{\begin{center}
    \tablestyle{4pt}{1.05}
    \begin{tabular}{y{72}x{24}x{24}}
    \shline
    case & top-1 & top-5 \\
    \shline
    L1 loss (w norm.)  &  81.51 & 95.58 \\
    L1 loss (w/o norm.) & 81.41 & 95.02 \\
    \baseline{MSE loss (w norm.)}  &  \baseline{\textbf{81.75}} & \baseline{\textbf{95.29}} \\
    MSE loss (w/o norm.) & 81.61 & 95.14 \\
    \end{tabular}
    \end{center}}\end{minipage}
    }
    \hspace{2em} 
    \subfloat[
    \textbf{ Number of TA blocks in  \textit{TATS}}. Our method performs best when \# of TA blocks $=$ 1.
    \label{tab:TATS_network}
    ]
    {
    \centering
    \begin{minipage}{0.36\linewidth}{\begin{center}
    \tablestyle{4pt}{1.05}
    \begin{tabular}{y{82}x{18}x{18}x{32}}
    \shline
    case & top-1 & top-5 & memory\\
    \shline
    \baseline{TA} $( \text{\# Blocks} = 1)$ & \baseline{\textbf{81.75}} & \baseline{\textbf{95.29}} & \baseline{\textbf{25.36 GB}} \\
    TA $(\text{\# Blocks} = 2)$ & 65.17 & 88.59 & 32.18 GB \\
    TA $(\text{\# Blocks} = 3)$ & 67.35 & 90.20 & 39.00 GB \\
    \end{tabular}
    \end{center}}\end{minipage}
     }
    \vspace{-.7em}
    \caption{Ablation analysis is conducted on the UCF101 dataset using models pre-trained with mask ratio $\rho=0.95$ for 400 epochs and fine-tuned on action recognition task for 100 epochs. The default choice of our method is highlighted in \colorbox{baselinecolor}{\bf gray} color.
    }
    \label{tab:ablation_table} 
    \vspace{-.99em}
\end{table*}



\noindent \textbf{2. Effect of Decoder Depth.~}Table~\ref{tab:decoder_depth} examines the impact of different decoder depths, specifically the number (\#) of transformer blocks in the decoder's architecture. Our findings show that the best performance is achieved with \# $\text{Blocks} = 1$, yielding a top-1 accuracy of $81.75\%$. This observation aligns with the results observed in \cite{bandara2023adamae,tong2022videomae}.

\noindent \textbf{3. Effect of Reconstruction Loss Function.}~In Table~\ref{tab:loss_target}, we examine the effect of the reconstruction objective, specifically comparing L1 and MSE losses. Following the standard approach introduced in VideoMAE~\cite{tong2022videomae}, we also explore computing these losses (L1/MSE) using both raw pixel values and per-patch normalized pixels. Our results indicate that MSE loss with per-patch normalization achieves the highest top-1 accuracy of 81.75\%.

\noindent \textbf{4. Effect of Number of Trajectory Attention Blocks. } In Table~\ref{tab:TATS_network}, we investigate the effect of varying the \# of TA blocks in \textit{TATS}. Our results indicate that the configuration with \# TA Blocks $=1$ yields the highest top-1 accuracy of $81.75\%$. As we increase the \# TA Blocks, the performance decreases while the memory usage increases.
 

\noindent \textbf{5. Memory Usage. } In Table~\ref{tab:memory_usage}, we inspect the memory usage of our approach in comparison to AdaMAE~\cite{bandara2023adamae} and VideoMAE~\cite{tong2022videomae}. Our method demonstrates lower memory consumption (pretraining) and better performance (finetuning) than AdaMAE~\cite{bandara2023adamae}. Although VideoMAE~\cite{tong2022videomae} utilizes less memory (pretraining) than our approach, our method significantly outperforms it in terms of top-1 accuracy (finetuning) on UCF101, achieving $81.75\%$ compared to only $65.86\%$ by VideoMAE~\cite{tong2022videomae}.


\section{Conclusions and Discussions}
\label{sec:conclusions}


In this work, we propose a novel and generalizable \textit{TATS} module that enhances MAE pre-training for videos by adaptively selecting motion-centric tokens based on their spatio-temporal motion trajectories. \textit{TATS} can be integrated into the MAE framework without requiring additional modalities like optical flow (e.g., RAFT~\cite{teed2020raft}) or external pre-trained models such as DINOv2~\cite{oquab2023dinov2} or CLIP~\cite{radford2021learning} for motion priors or semantic cues. We also introduce a unified training framework (Algorithm~\ref{ppo_training_recipe}) that enables the joint optimization of MAE and \textit{TATS} from scratch using PPO~\cite{schulman2017proximal}, enhancing stability during pre-training even under aggressive masking. Finally, we perform an extensive quantitative, qualitative and ablation assessment (Tables~\ref{finetuning_results},\ref{transfer_results},\ref{tab:ablation_table}) on benchmark datasets (K400, SSv2, UCF101, HMDB51) for the downstream task of action recognition, showcasing the effectiveness, generalization, transferability, and efficiency of our approach compared to state-of-the-art methods.

\noindent \textbf{Future Work.} Our proposed \textit{TATS} and training recipe does need to be empirically validated on other downstream tasks and extended to other modalities. Furthermore, with the recent resurgence in RL research due to its applications in LLMs, it is important to reconsider strategies that integrate dynamic computation into masked modeling approaches, optimizing them through RL algorithms.   We plan to conduct future studies around these topics. We hope this work can motivate further research in this direction.

%
\section{Acknowledgement}
This research is supported by Research Ireland under the Grant Number SFI/12/RC/2289\_P2 and Irish Centre for High End Computing (ICHEC).

{
    \small
    \bibliographystyle{ieeenat_fullname}
    \bibliography{main}
}

\clearpage
\setcounter{page}{1}
\maketitlesupplementary


\section{Datasets}

\noindent \textbf{Something-Something V2} (SSv2) \cite{goyal2017something} is a curated video dataset for human action classification, comprising 174 classes and a total of 220,847 videos. Each video depicts a single action with a duration ranging from 2 to 6 seconds. SSv2 is a motion-focused dataset, where temporal relationships are more pronounced compared to other datasets.

\noindent \textbf{Kinetics-400} (K400) \cite{kay2017kinetics} is a widely used large-scale video dataset, comprising 400 classes and over 250,000 videos. Each video, approximately 10 seconds in duration, captures a single action.

\noindent \textbf{HMDB51} \cite{kuehne2011hmdb} comprises 51 classes and a total of 6,766 videos. HMDB51 emphasizes appearance information over motion dynamics.

\noindent \textbf{UCF101} \cite{soomro2012ucf101} comprises 13,320 video clips categorized into 101 classes. These classes span five activity types: body motion, human-to-human interaction, human-to-object interaction, musical instrument performance, and sports.

\begin{table}[b]
    \centering
    \small
    \caption{Hyperparameter setting for pre-training across all benchmark datasets.}  
    \label{tab: hyper-param_pretrain}
    \begin{tabular}{l | c } 
        \toprule
        Configuration & Value\\  
        \hline
       Learning rate for $g_\theta$ - $lp$  & 1.5e-6  \\  
       Epochs to train $f_\phi$ only - $m_o$ & 10 \\
       Steps to train $f_\phi$ and record $g_\theta$ episodes - $k$ & 1 \\
       Softmax Temperature & 1 \\
       Policy loss coefficient - $c_1$ & 1e-4 \\
       Value loss coefficient - $c_2$ & 1e-4 \\
       Entropy coefficient - $c_3$ & 1e-4 \\    
       Optimizer & AdamW \\
       Optimizer betas & {0.9, 0.95} \\
       Batch size & 32 \\
       Base learning rate & 1.5e-4 \\
       Learning rate schedule & cosine decay \\
       Warmup epochs & 40 \\
       Augmentation & MultiScaleCrop \\
        \bottomrule
        \hline
    \end{tabular}
\end{table}

\section{Additional Implementation Details}

\subsection{Data Preprocessing}

Our data processing pipeline closely follows AdaMAE~\cite{bandara2023adamae} for pre-training. We extract 16 frames of dimension $224 \times 224$ from the videos, using a temporal stride of 4 (K400) and 2 (HMDB51/UCF101/SSv2), with the starting frame randomly selected~\cite{feichtenhofer2022masked}. During pre-training, we apply data augmentation techniques, including random resized cropping in the spatial domain, random scaling within the range 
$ \in [ 0.5, 1 ]$, and random horizontal flipping~\cite{feichtenhofer2022masked}.

\subsection{Hyper-parameter Setting}

\noindent \textbf{Pre-training.}  The hyperparameter configurations used during the pre-training phase across all benchmark datasets are presented in Table~\ref{tab: hyper-param_pretrain}. For $(m_o, k)$, hyperparameter tuning is conducted on the UCF101 and HMDB51 datasets (Table~\ref{tab:hyper-param_tuning_pretraining_m_o_k}), and the configuration that minimizes the reconstruction error is selected. Similarly we also perform hyperparameter tuning for coefficients ($c_1$,$c_2$,$c_3$) in Table~\ref{tab:hyper-param_tuning_pretraining_coefficients} during pretraining on UCF101 and observe that (1e-4, 1e-4, 1e-4) minimizes the reconstruction error. Empirical observations indicate that the optimal configuration for UCF101 also performs effectively on subset of K400 and SSv2 (\underline{small scale pre-training setup}). It is to be noted that we use reconstruction loss for tuning these hyper-parameters because behaviour of reconstruction loss during pretraining is more interpretable in terms of convergence than the sampling loss.



\noindent \textbf{Fine-tuning. } The hyperparameter setting for end-to-end fine-tuning on the downstream task of action recognition across all benchmarks is summarized in Table~\ref{tab: hyper-param_finetune}.



\begin{table}[t]
    \centering
    \small
    \caption{Hyperparameter ($m_o,k$) tuning for pre-training, evaluated based on reconstruction error on UCF101 and HMDB51. Same configuration is adopted for SSv2 and K400 as in UCF101.}  
    \label{tab:hyper-param_tuning_pretraining_m_o_k}
    \begin{tabular}{l | c c } 
        \toprule
        ($m_o$, $k$) & UCF101 & HMDB51 \\  
        \hline
        \baseline{(0, 1)} & 0.5211 & \baseline{0.8051} \\ 
        (1, 1) & 0.5205 & 0.8195 \\
        (5, 1) & 0.5304 & 0.8535 \\
        \baseline{(10, 1)} & \baseline{0.5135} & 0.8278 \\
        (25, 1) & 0.5269 & 0.8987 \\
        (100, 1) & 0.6662 & 0.9291 \\
        (50, 5) & 0.7735 &  0.9772 \\
        (50, 10) & 0.8149 & 0.9776 \\
        (50, 25) & 0.9201 & - \\
        
        \bottomrule
        \hline
    \end{tabular}
\end{table}

\begin{table}[h]
    \centering
    \small
    \caption{Hyperparameter ($c_1,c_2,c_3$) tuning for pre-training, evaluated based on reconstruction error on UCF101. Same configuration is adopted for SSv2, K400 and HMDB51. $(m_o,k)$ are fixed as $(10,1)$}  
    \label{tab:hyper-param_tuning_pretraining_coefficients}
    \begin{tabular}{l | c } 
        \toprule
        ($c_1$, $c_2$, $c_3$) & UCF101 \\  
        \hline
        (1e-4, 1e-3, 1e-3) & 0.5188 \\
        (1e-4, 1e-3, 1e-4) & 0.5167 \\
        (1e-4, 1e-4, 1e-3) & 0.5246 \\
        (1e-3, 1e-4, 1e-4) & 0.8482 \\
        \baseline{(1e-4, 1e-4, 1e-4)} & \baseline{0.5135} \\
        (1e-5, 1e-4, 1e-4) & 0.5239 \\
        (1e-3,1e-3,1e-4) & 0.5215 \\
        (1e-3, 1e-3, 1e-4) & 0.7869 \\
        (1e-5, 1e-5, 1e-5) & 0.5173 \\
        \bottomrule
        \hline
    \end{tabular}
\end{table}


\begin{table}[h]
    \centering
    \small
    \caption{Hyperparameter setting for end-to-end fine-tuning for all benchmark datasets.}  
    \label{tab: hyper-param_finetune}
    \begin{tabular}{l | c } 
        \toprule
        Configuration & Value \\  
        \hline
        Optimizer & AdamW \\
        Optimizer Betas & \{0.9, 0.999\} \\
        Batch size & 8 \\
        Weight Decay & 5e-2 \\
        Base Learning Rate & 1e-3 \\
        Learning Rate Schedule & cosine decay \\
        Layer-wise learning rate decay & 0.75 \\
        Warmup epochs & 5 \\ 
        RandAug & \(9,0.5\) \\
        Label Smoothing & 0.1 \\
        Mixup & 0.8 \\
        DropPath & 0.1\\
        \# Temporal Clips & 5 \small (k400), 2 (ssv2/hmdb/ucf) \\
        \# Spatial Crops & 3 \\
        \bottomrule
        \hline
    \end{tabular}
\end{table}

\begin{table}[t]
    \centering
    \tablestyle{2.0pt}{1.04}
    \caption{Encoder-Decoder architecture based on AdaMAE~\cite{bandara2023adamae}. TATS : Trajectory Aware Adaptive Token Sampler. MHA : Multi-Head Self-Attention}
    \begin{tabular}{lcc}
    \hline
    Stage & ViT-Base & Output shape \\
    \hline
    \multirow{2}{*}{Input Video} & stride $\textcolor{blue}{4} \times \textcolor{darkgreen}{1} \times \textcolor{darkgreen}{1}$ for K400 & \multirow{2}{*}{$\textcolor{red}{3} \times \textcolor{blue}{16} \times \textcolor{darkgreen}{224} \times \textcolor{darkgreen}{224}$}\\
    & stride $\textcolor{blue}{2} \times \textcolor{darkgreen}{1} \times \textcolor{darkgreen}{1}$ for ssv2/ucf/hmdb & \\
    \hline
    \multirow{2}{*}{Tokenization} &  stride $\textcolor{blue}{2} \times \textcolor{darkgreen}{16} \times \textcolor{darkgreen}{16}$ & \multirow{2}{*}{$\textcolor{brown}{1568} \times \textcolor{red}{768} $}\\
    & emb. dim \textcolor{red}{768} & \\
    &  kernel size $\textcolor{blue}{2} \times \textcolor{darkgreen}{16} \times \textcolor{darkgreen}{16}$ & \\
    \hline
    
    \multirow{2}{*}{Masking} & TATS Masking & \multirow{2}{*}{$[(1-\rho) \times \textcolor{brown}{1568}] \times \textcolor{red}{768} $} \\
    & mask ratio $\rho$ & \\
    \hline
    
    \multirow{2}{*}{Encoder} & \multirow{2}{*}{$[MHA(\textcolor{red}{768})]\times 12$} &\multirow{2}{*}{$[(1-\rho) \times \textcolor{brown}{1568}] \times \textcolor{red}{768}$} \\ 
    & &\\
    \hline
    
    \multirow{2}{*}{Projection} & $MHA(\textcolor{red}{384})$ &\multirow{2}{*}{$ \textcolor{brown}{1568} \times \textcolor{red}{384}$} \\ 
    & concat masked tokens &\\
    \hline
    
    \multirow{2}{*}{Decoder} & \multirow{2}{*}{$[MHA(\textcolor{red}{384})]\times 4$} &\multirow{2}{*}{$[(1-\rho) \times \textcolor{brown}{1568}] \times \textcolor{red}{384}$} \\ 
    & &\\
    \hline
    
    \multirow{2}{*}{Projector} & \multirow{2}{*}{$MLP(\textcolor{red}{1536})$} &\multirow{2}{*}{$\textcolor{brown}{1568} \times \textcolor{red}{1536}$} \\ 
    &  &\\
    \hline
    
    \multirow{2}{*}{Reshaping} & \multirow{2}{*}{from \textcolor{red}{1536} to $\textcolor{red}{3} \times \textcolor{blue}{2} \times \textcolor{darkgreen}{16} \times \textcolor{darkgreen}{16}$} &\multirow{2}{*}{$\textcolor{red}{3} \times \textcolor{blue}{16} \times \textcolor{darkgreen}{224} \times \textcolor{darkgreen}{224}$} \\ 
    &  &\\
    \hline
    \end{tabular}
    \label{tab:encoder_decoder_arch}
\end{table}

\subsection{Encoder-Decoder Architecture}




We adopt an asymmetric encoder-decoder architecture~\cite{bandara2023adamae} for self-supervised pre-training and augment it with \textit{TATS} module and only keep the encoder during the fine-tuning. In particular, the design of the encoder-decoder is based on 16-frame vanilla ViT-Base architecure. Table \ref{tab:encoder_decoder_arch} provides an overview of the encoder-decoder architecture utilized in our framework. 

\section{Large Scale Pre-training Results}
We conduct pre-training (400 epochs) and finetuning (100 epochs) of our model on full SSv2~\cite{goyal2017something} dataset for $\rho=0.95$ on $8$ Nvidia A100 GPUs. In order to ensure fairness in comparison, we also pre-train (400 epochs) and finetune (100 epochs) both baselines VideoMAE~\cite{tong2022videomae} and AdaMAE~\cite{bandara2023adamae} on full SSv2 for $\rho=0.95$ with the same GPU setup using their public source code and default configuration.

Table~\ref{large_scale_finetune_ssv2} presents the top-1 and top-5 accuracy obtained in this experiment. We observe that our approach outperforms both the baselines under aggressive masking setting even for large scale experiments. This highlights the effectiveness and generalization capability of the proposed \textit{TATS} module and the training strategy in terms of learning a better feature quality than learnt by \cite{bandara2023adamae,tong2022videomae}.

Due to the availability of limited computational resources, our experiments in this setup is limited.

\begin{table}[t]
    \centering
    \small
    \caption{\small \textbf{Large Scale Pre-training and Finetuning Results.} Comparison of fine-tuning result of \colorbox{baselinecolor}{ Our} model against baselines~(\cite{bandara2023adamae,tong2022videomae}) on action recognition task for full SSv2 and $\rho=0.95$ with top-1/top-5 accuracy as evaluation metric. (\up{} : denotes increase in performance)}
    \label{large_scale_finetune_ssv2}
    \begin{tabular}{lll} 
        \toprule 
        Method & top-1 & top-5  \\ 
        \hline 
        VideoMAE~\cite{tong2022videomae}$_{\rho=95\%}$   & 59.38  & 84.17   \\ 
        AdaMAE~\cite{bandara2023adamae}$_{\rho=95\%}$   & 63.06   & 85.89   \\ 
        \baseline{Ours$_{\rho=95\%}$}   & \baseline{65.82} \up{}  & \baseline{88.50}\up{}  \\ 
        \bottomrule
    \end{tabular}
    \label{tab:example}
\end{table}

\section{Mask Visualization}
\begin{figure}[t]
    \centering
    \includegraphics[width=0.5\textwidth]{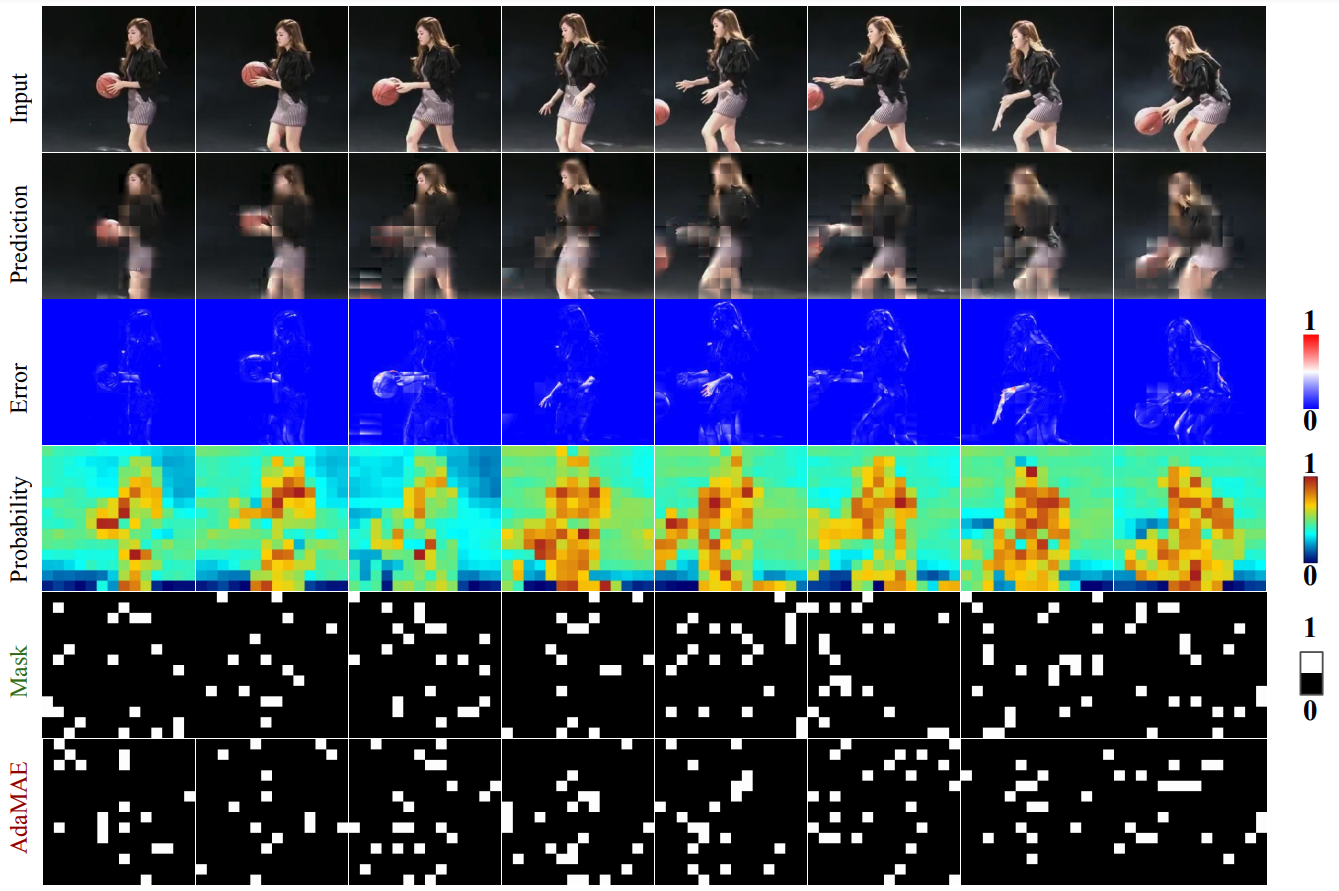}
    \caption{Sample Visualization of a Kinetics 400 video with \textcolor{darkgreen}{adaptive sampling using \textit{TATS}} with mask ratio $\rho=0.95$. Compared with \textcolor{darkbrown}{AdaMAE}~\cite{bandara2023adamae} masks.}
    \label{fig:sample_k400_ours_0.95}
\end{figure}

\begin{figure}[t]
    \centering
    \includegraphics[width=0.5\textwidth]{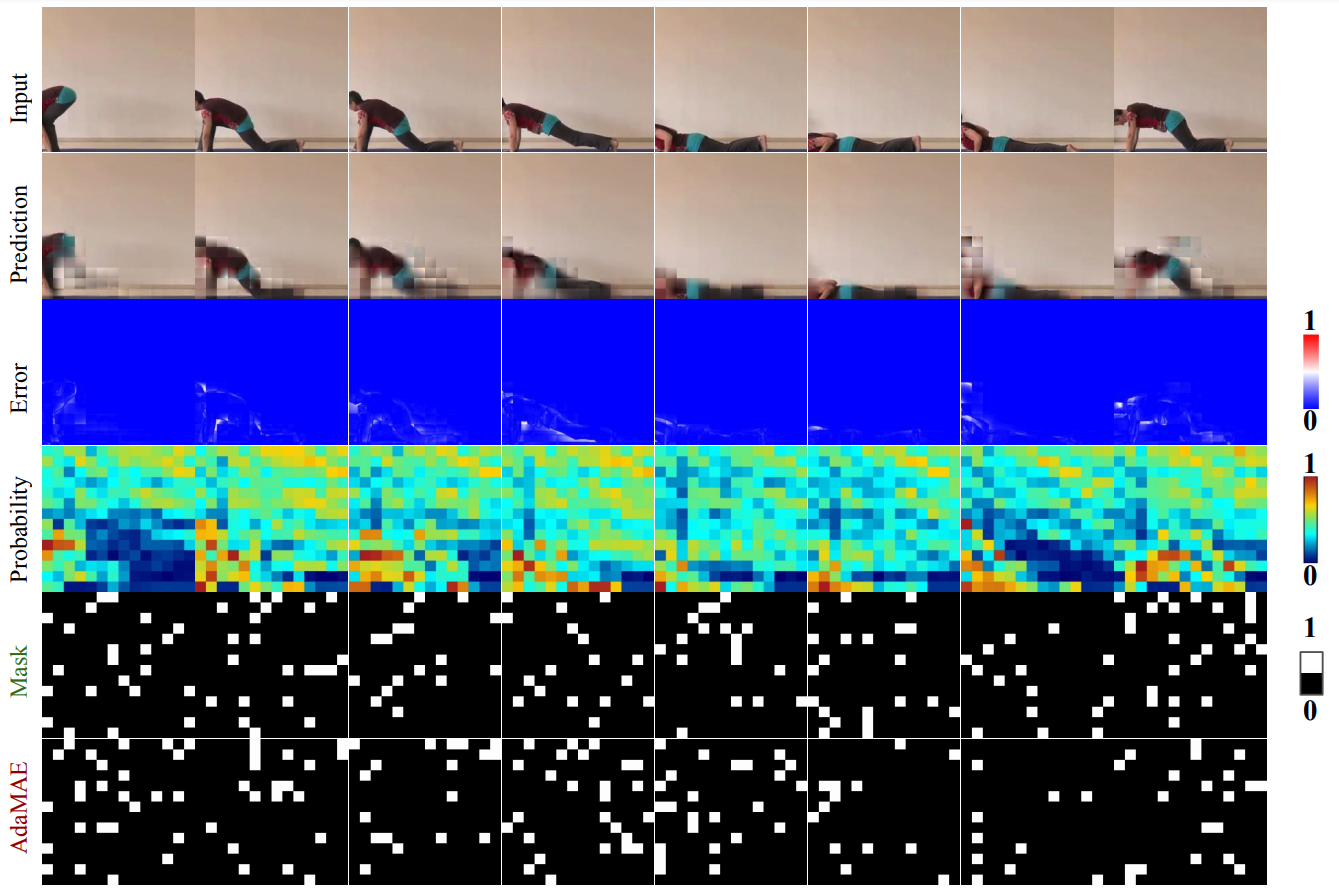}
    \caption{Sample Visualization of a Kinetics 400 video with \textcolor{darkgreen}{adaptive sampling using \textit{TATS}} with mask ratio $\rho=0.9$. Compared with \textcolor{darkbrown}{AdaMAE}~\cite{bandara2023adamae} masks.}
    \label{fig:sample_k400_ours_0.9}
\end{figure}

\begin{figure}[t]
    \centering
    \includegraphics[width=0.5\textwidth]{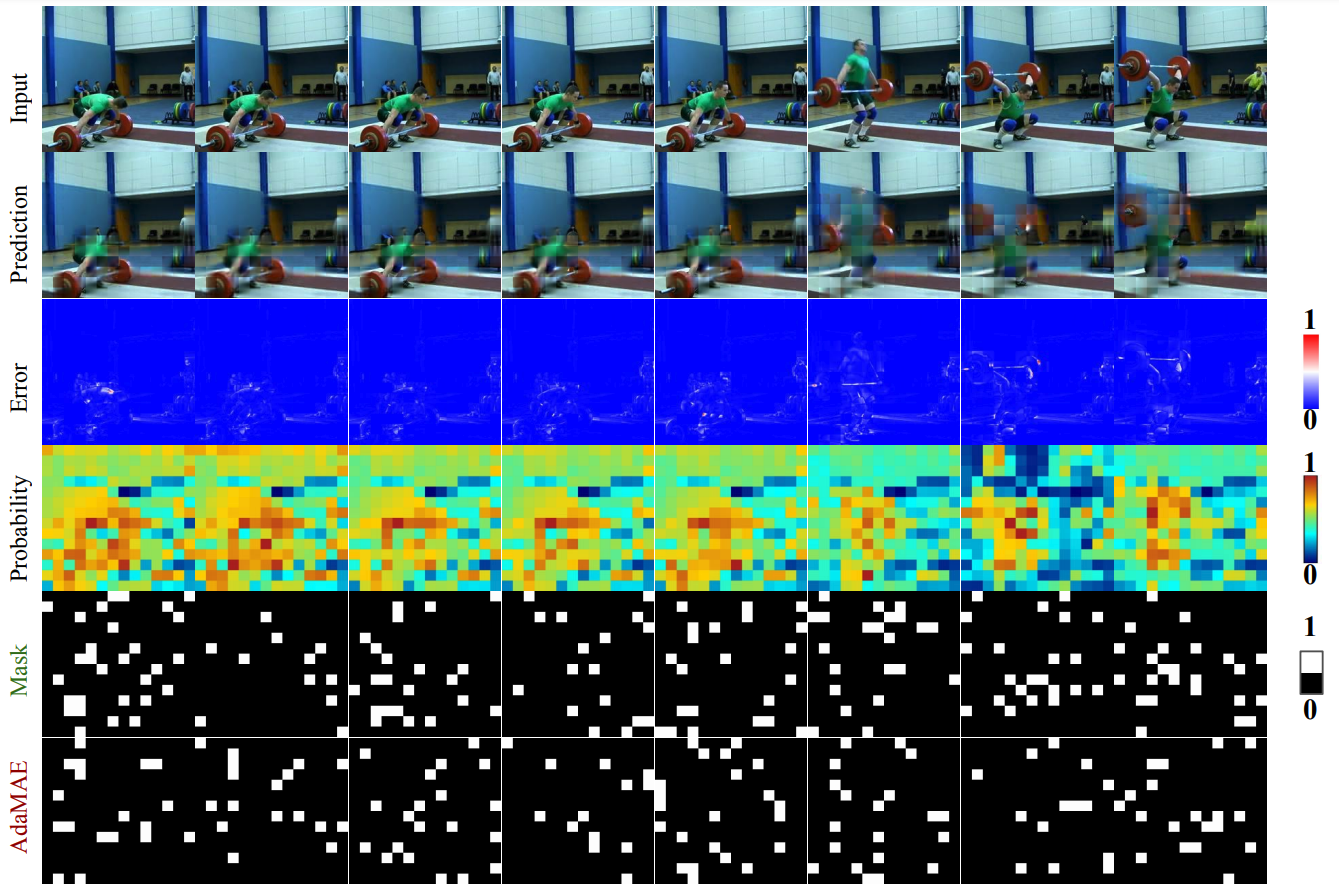}
    \caption{Sample Visualization of a Kinetics 400 video with \textcolor{darkgreen}{adaptive sampling using \textit{TATS}} with mask ratio $\rho=0.85$. Compared with \textcolor{darkbrown}{AdaMAE}~\cite{bandara2023adamae} masks.}
    \label{fig:sample_k400_ours_0.85}
\end{figure}

\begin{figure}[h]
    \centering
    \includegraphics[width=0.5\textwidth]{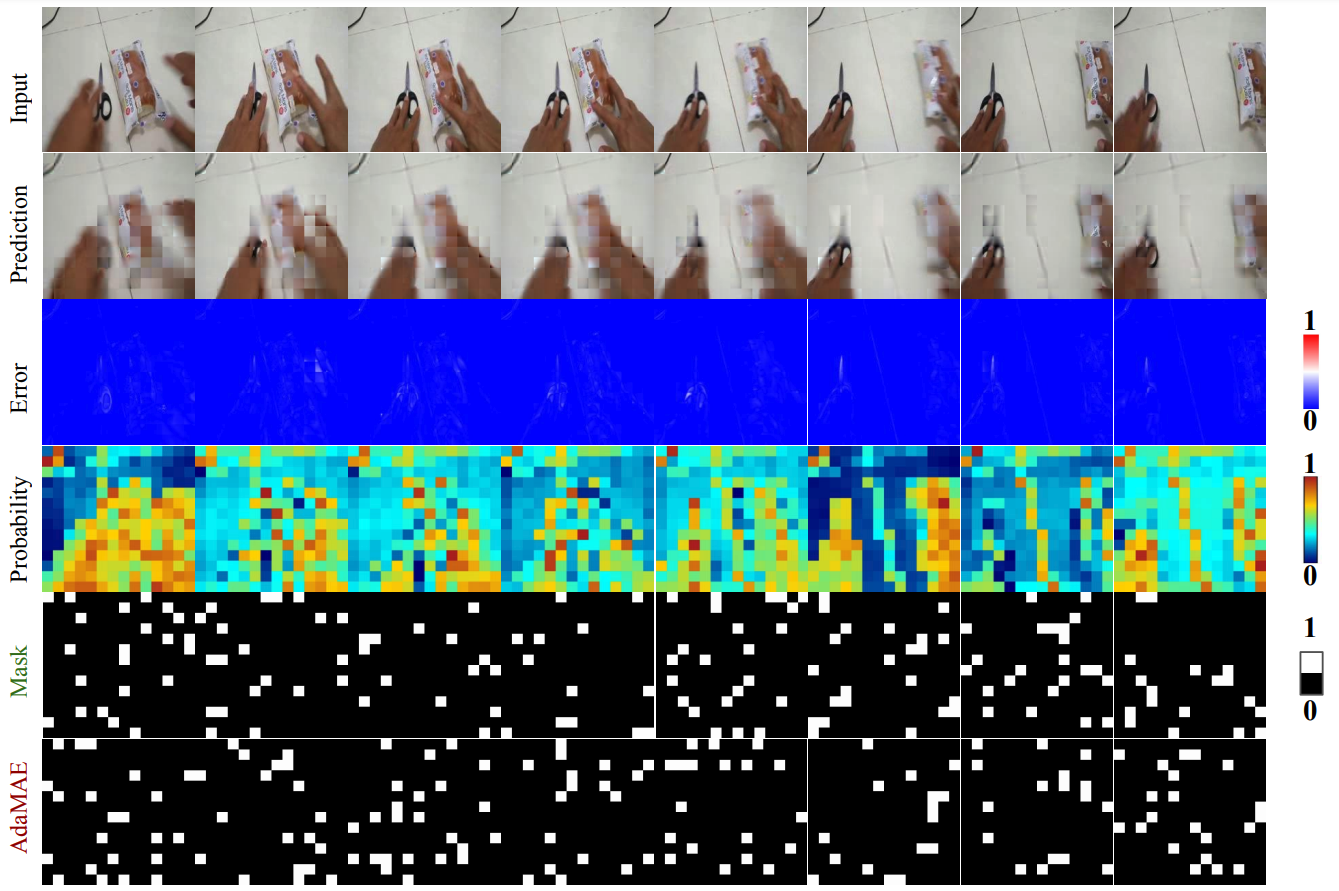}
    \caption{Sample Visualization of a SSv2 video with \textcolor{darkgreen}{adaptive sampling using \textit{TATS}} with mask ratio $\rho=0.95$. Compared with \textcolor{darkbrown}{AdaMAE}~\cite{bandara2023adamae} masks.}
    \label{fig:sample_ssv2_ours_0.95}
\end{figure}

\begin{figure}[h]
    \centering
    \includegraphics[width=0.5\textwidth]{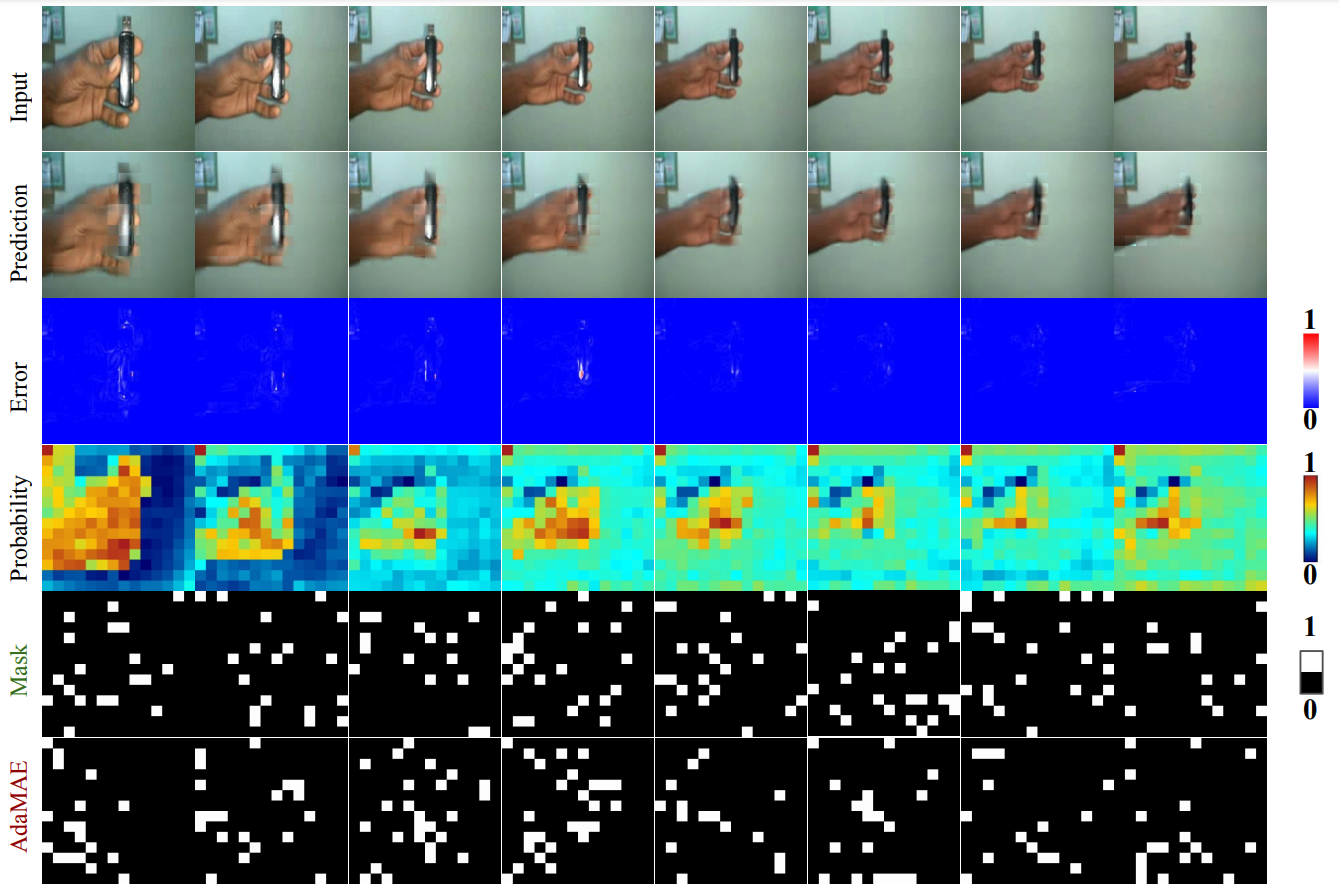}
    \caption{Sample Visualization of a SSv2 video with \textcolor{darkgreen}{adaptive sampling using \textit{TATS}} with mask ratio $\rho=0.9$. Compared with \textcolor{darkbrown}{AdaMAE}~\cite{bandara2023adamae} masks.}
    \label{fig:sample_ssv2_ours_0.9}
\end{figure}

\begin{figure}[h]
    \centering
    \includegraphics[width=0.5\textwidth]{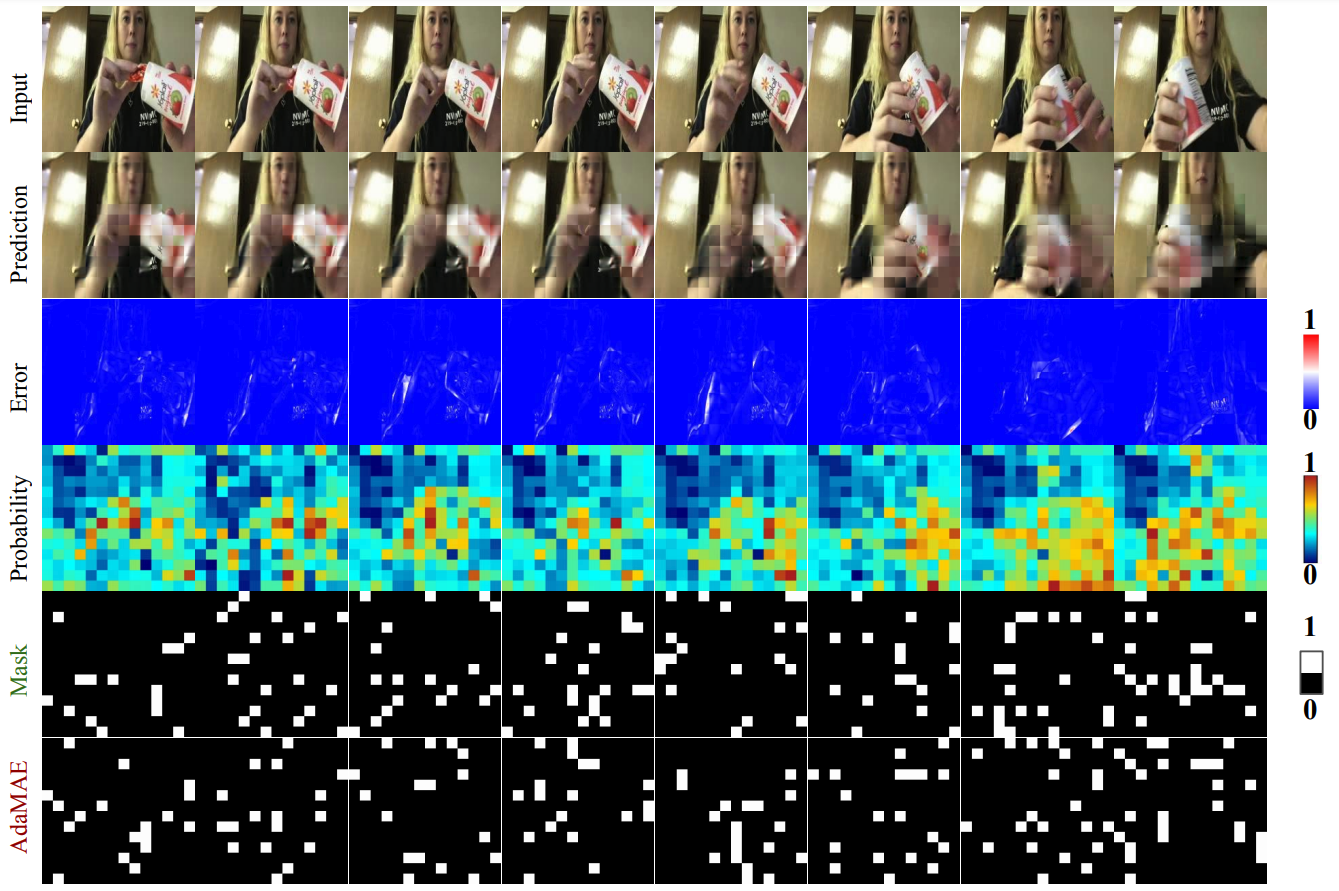}
    \caption{Sample Visualization of a SSv2 video with \textcolor{darkgreen}{adaptive sampling using \textit{TATS}} with mask ratio $\rho=0.85$. Compared with \textcolor{darkbrown}{AdaMAE}~\cite{bandara2023adamae} masks.}
    \label{fig:sample_ssv2_ours_0.85}
\end{figure}

\begin{figure}[h]
    \centering
    \includegraphics[width=0.5\textwidth]{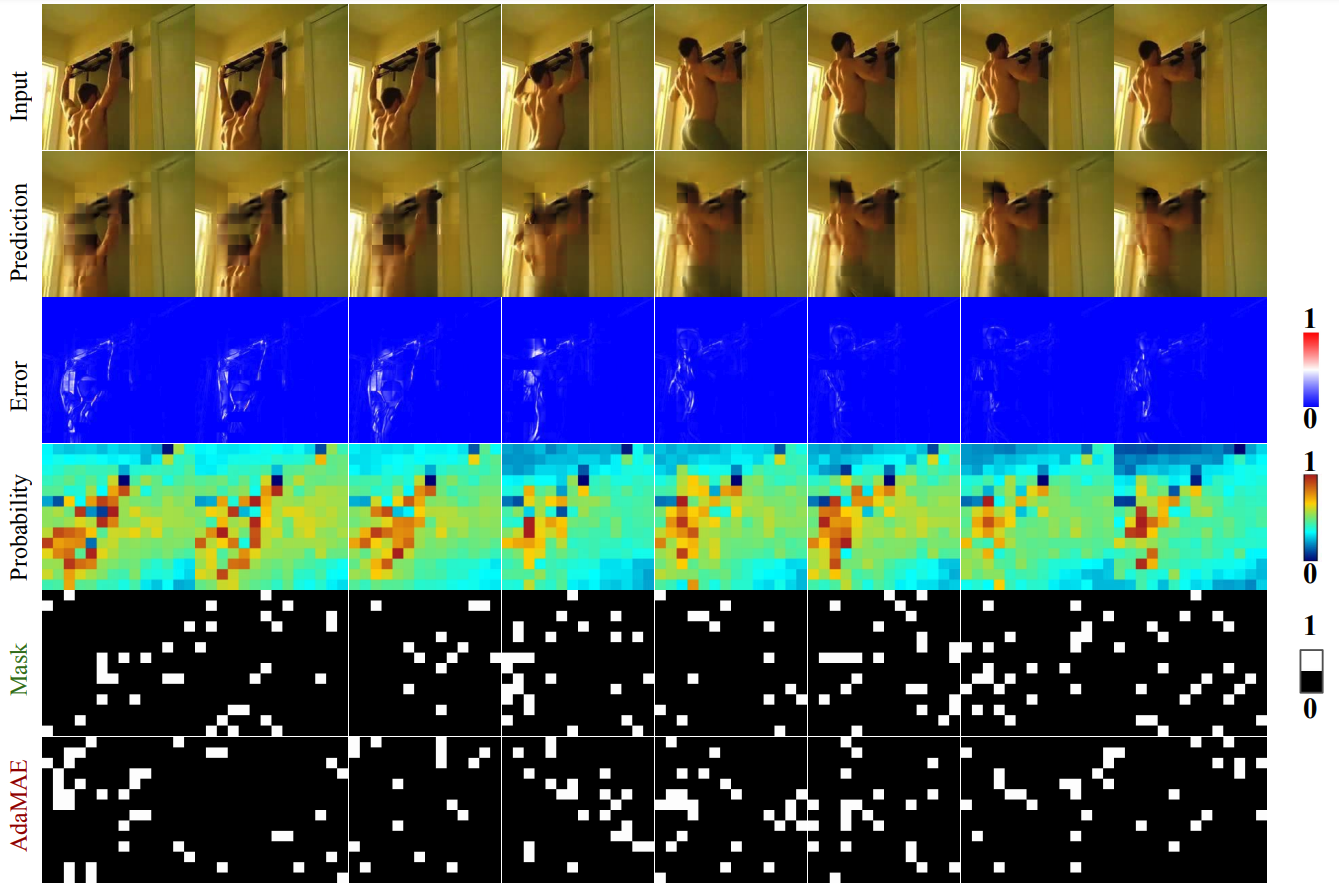}
    \caption{Sample Visualization of a UCF101 video with \textcolor{darkgreen}{adaptive sampling using \textit{TATS}} with mask ratio $\rho=0.95$. Compared with \textcolor{darkbrown}{AdaMAE}~\cite{bandara2023adamae} masks.}
    \label{fig:sample_ucf_ours_0.95}
\end{figure}

\begin{figure}[h]
    \centering
    \includegraphics[width=0.5\textwidth]{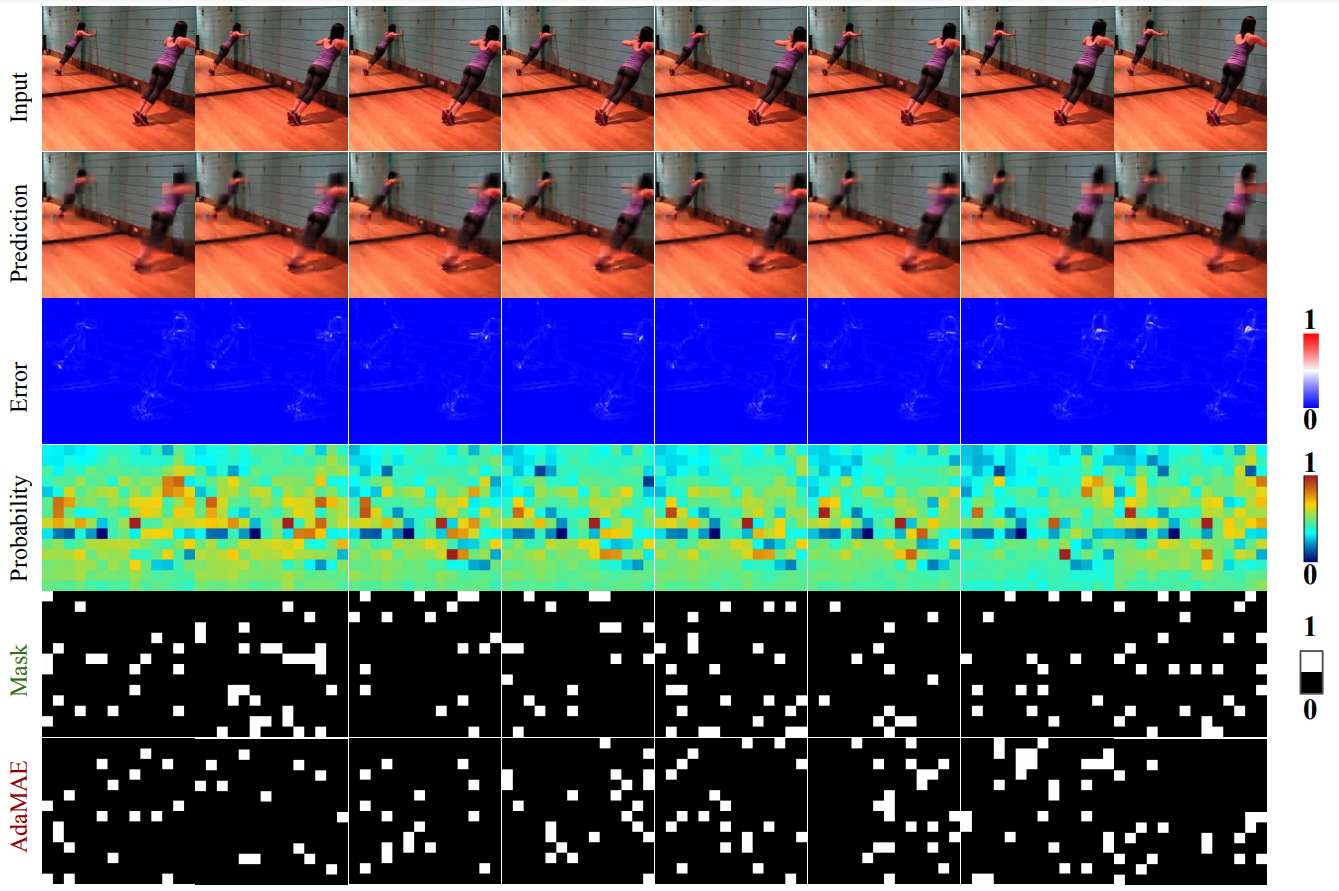}
    \caption{Sample Visualization of a UCF101 video with \textcolor{darkgreen}{adaptive sampling using \textit{TATS}} with mask ratio $\rho=0.9$. Compared with \textcolor{darkbrown}{AdaMAE}~\cite{bandara2023adamae} masks.}
    \label{fig:sample_ucf_ours_0.9}
\end{figure}

\begin{figure}[h]
    \centering
    \includegraphics[width=0.5\textwidth]{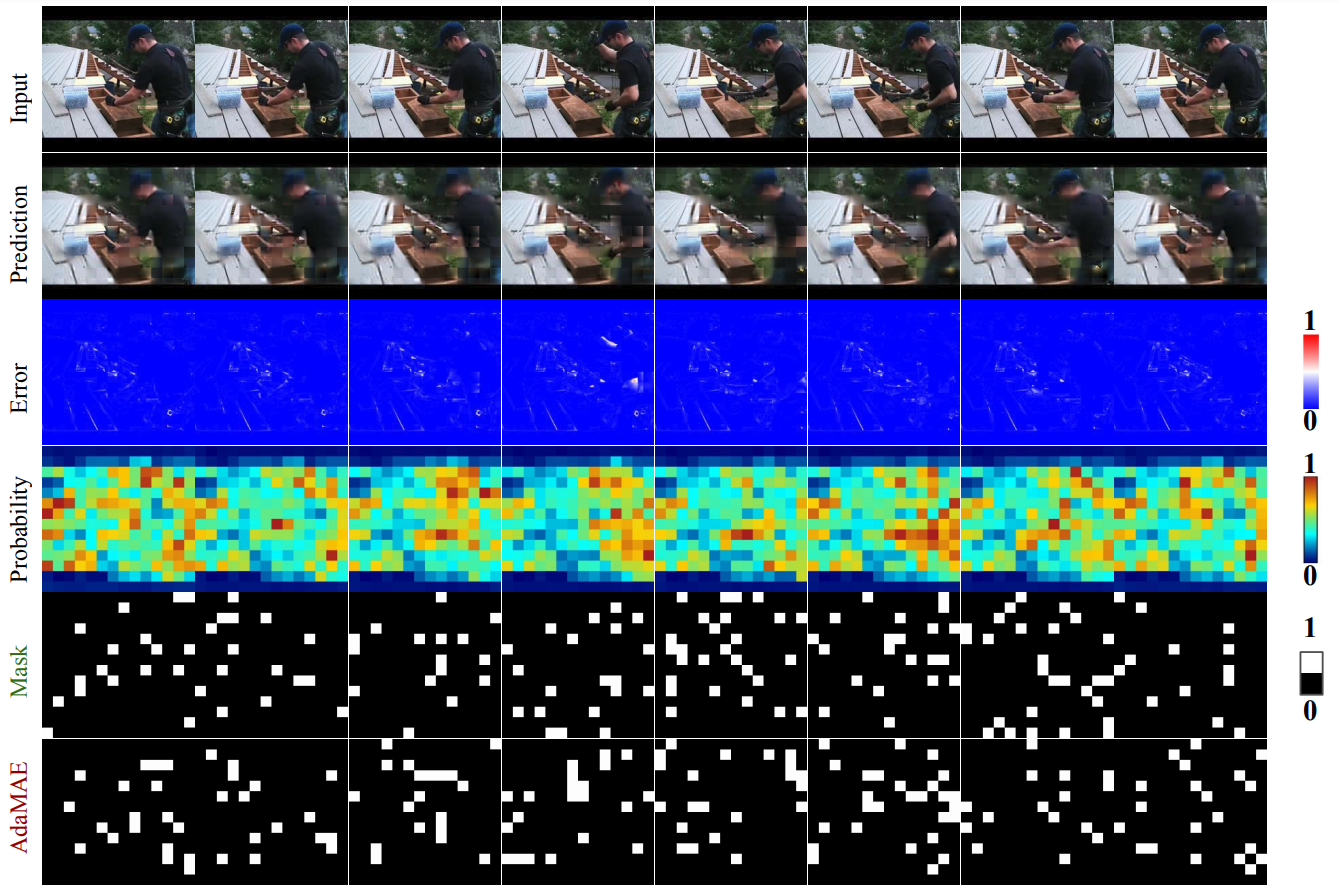}
    \caption{Sample Visualization of a UCF101 video with \textcolor{darkgreen}{adaptive sampling using \textit{TATS}} with mask ratio $\rho=0.85$. Compared with \textcolor{darkbrown}{AdaMAE}~\cite{bandara2023adamae} masks.}
    \label{fig:sample_ucf_ours_0.85}
\end{figure}

\begin{figure}[h]
    \centering
    \includegraphics[width=0.5\textwidth]{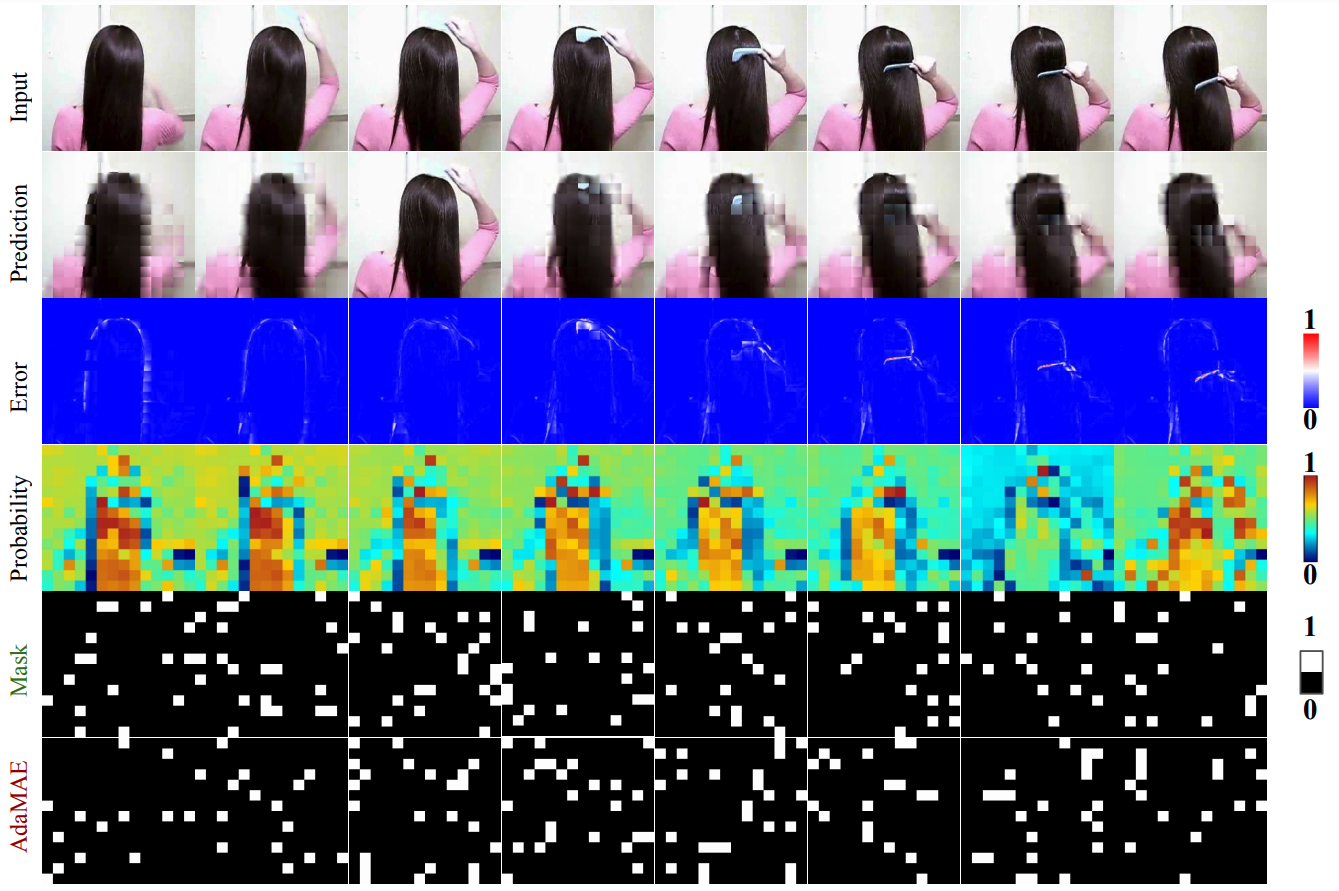}
    \caption{Sample Visualization of a HMDB51 video with \textcolor{darkgreen}{adaptive sampling using \textit{TATS}} with mask ratio $\rho=0.95$. Compared with \textcolor{darkbrown}{AdaMAE}~\cite{bandara2023adamae} masks.}
    \label{fig:sample_hmdb_ours_0.95}
\end{figure}

\begin{figure}[h]
    \centering
    \includegraphics[width=0.5\textwidth]{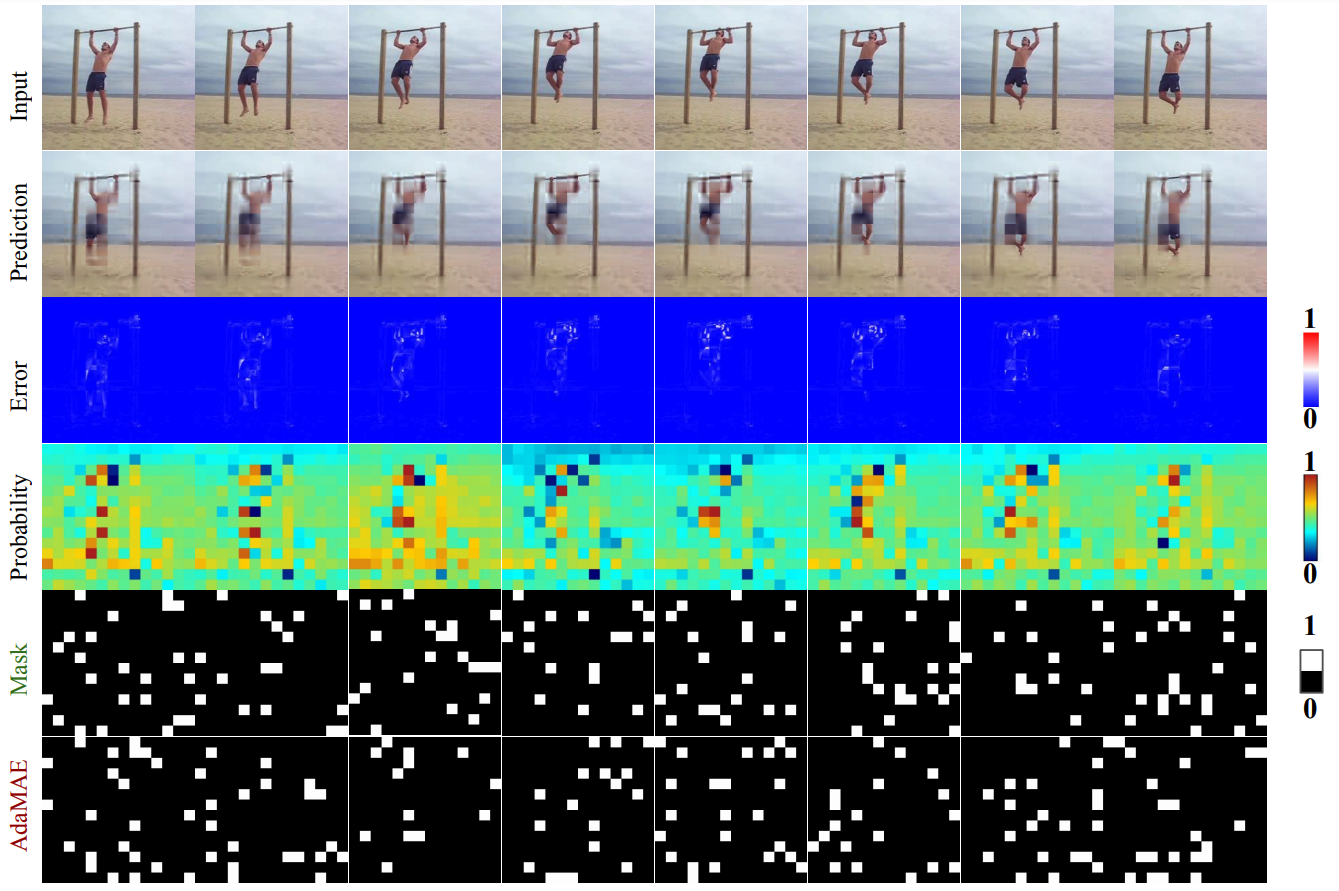}
    \caption{Sample Visualization of a HMDB51 video with \textcolor{darkgreen}{adaptive sampling using \textit{TATS}} with mask ratio $\rho=0.9$. Compared with \textcolor{darkbrown}{AdaMAE}~\cite{bandara2023adamae} masks.}
    \label{fig:sample_hmdb_ours_0.9}
\end{figure}

\begin{figure}[h]
    \centering
    \includegraphics[width=0.5\textwidth]{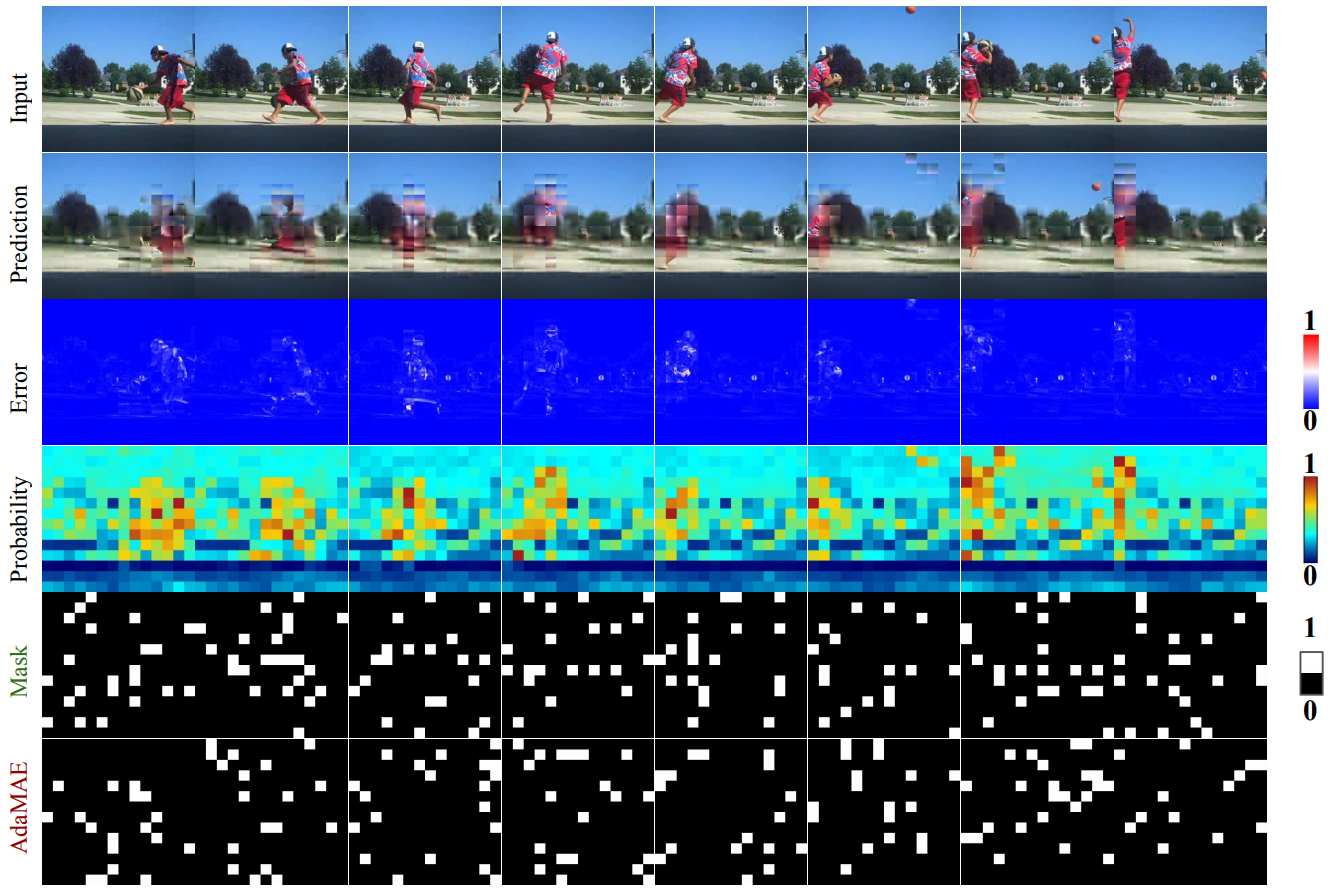}
    \caption{Sample Visualization of a HMDB51 video with \textcolor{darkgreen}{adaptive sampling using \textit{TATS}} with mask ratio $\rho=0.85$. Compared with \textcolor{darkbrown}{AdaMAE}~\cite{bandara2023adamae} masks.}
    \label{fig:sample_hmdb_ours_0.85}
\end{figure}

Here we show visualizations \textcolor{darkgreen}{adaptive sampling learned by our \textit{TATS} module} across benchmark dataset for different mask ratios $\rho = \{0.95, 0.9, 0.85\}$ in Figure~\ref{fig:sample_k400_ours_0.95}, \ref{fig:sample_k400_ours_0.9}, \ref{fig:sample_k400_ours_0.85},  \ref{fig:sample_ssv2_ours_0.95}, \ref{fig:sample_ssv2_ours_0.9}, \ref{fig:sample_ssv2_ours_0.85}, \ref{fig:sample_ucf_ours_0.95}, \ref{fig:sample_ucf_ours_0.9}, \ref{fig:sample_ucf_ours_0.85}, \ref{fig:sample_hmdb_ours_0.95}, \ref{fig:sample_hmdb_ours_0.9}, \ref{fig:sample_hmdb_ours_0.85}.

In all of these Figures, \underline{first row} represents input video frames, the \underline{second row} depicts the prediction/reconstruction, the \underline{third row} shows the reconstruction error, the \underline{fourth row} represents the probability of sampling the space-time patch, \underline{fifth row} shows the \textcolor{darkgreen}{adaptive masks learned by  \textit{TATS}}. The \underline{last row} depicts the binary masks learned by \textcolor{darkbrown}{AdaMAE}~\cite{bandara2023adamae} for comparison.

\end{document}